\documentclass[reqno,11pt]{amsart}
\usepackage{amsmath,amsthm,amsfonts,color,graphicx,bbm,mathabx}

\usepackage[latin1]{inputenc}
\usepackage[makeroom]{cancel}
\usepackage{pdfsync,subfigure,multirow,float,esint}
\usepackage{filecontents,diagbox,bbold,dsfont}
\usepackage{microtype}

\definecolor{lightGray}{RGB}{235,235,235}
\definecolor{orange}{RGB}{255,128,0}
\definecolor{ucib}{RGB}{0,36,105}
\definecolor{mygreen}{RGB}{0,128,0}
\definecolor{lightBlue}{RGB}{102,153,204}

\newtheorem{thm}{Theorem}[section]
\newtheorem{lem}[thm]{Lemma}

\newtheorem{prop}[thm]{Proposition}

\newtheorem{rem}[thm]{Remark}

\DeclareMathAlphabet{\mathpzc}{OT1}{pzc}{m}{it}

\numberwithin{equation}{section}
\numberwithin{figure}{subsection}
\numberwithin{table}{subsection}

\begin{document}
\bibliographystyle{plain}

\title{Extracting Manifold Information from Point Clouds}

\author{Patrick Guidotti}
\address{University of California, Irvine\\
Department of Mathematics\\
340 Rowland Hall\\
Irvine, CA 92697-3875\\ USA }
\email{gpatrick@math.uci.edu}

\begin{abstract}
A kernel based method is proposed for the construction of signature
(defining) functions of subsets of $\mathbb{R}^d$. The subsets can
range from full dimensional manifolds (open subsets) to point clouds
(a finite number of points) and include bounded (closed) smooth manifolds of
any codimension. The interpolation and analysis of point clouds are
the main application. Two extreme cases in terms of regularity are
considered, where the data set is interpolated by an analytic surface,
at the one extreme, and by a H\"older continuous surface, at the
other. The signature function can be computed as a combination of
translated kernels, the coefficients of which are the solution of a 
Fredholm integral equation (matrix equation in the finite dimensional
case). Once it is obtained, it can be used to estimate the dimension
as well as the normal and the curvatures of the interpolated
manifold. The method is global and does not require 
the data set to be organized or structured in any particular way. It
admits a variational formulation with a natural regularized
counterpart, that proves useful in dealing with data sets corrupted by
numerical error or noise. The underlying analytical structure of the
approach is presented in general before it is applied to the case of
point clouds.
\end{abstract}

\subjclass[1991]{}

\maketitle

\section{Introduction}
The main goal of this paper is to propose a method to compute
geometric information about a manifold (a hypersurface, in many cases)
that is merely given as a point cloud, i.e. a set of points that are
assumed to be a sampling of the given manifold. While this may not
always be the primary goal of manifold learning, 
the connection to it is obvious and warrants some discussion. Manifold
learning is mainly motivated by the desire to find a ``simpler'' lower
dimensional representation for data that, while embedded in a space of
high dimension, are intrinsically of lower dimension. This is clearly
the case when the data represent a discrete sample of a low dimensional
manifold sitting in a high dimensional (Euclidean, for most purposes)
space, but the ideas and techniques are used well beyond this ``pure''
setting in applications. The first attempts were historically in a
(fundamentally) linear context and the core ideas were those of Principal Component
Analysis (PCA) \cite{P1901,H1933} and Multidimensional Scaling (MDS)
\cite{T1952,G1966}. The first attempts to find orthogonal axes along
which most of the variance in the data is captured, whereas the second
looks for a lower dimensional embedding that preserves the distances
between the data points. In many applications of interest, however,
data do not lie on a linear manifold (even in the absence of noise)
and points that are close to each other in the ambient space may be
far from each other on the manifold. This led to the development of
methods capable of detecting the nonlinear nature of the (data)
manifold. The idea is often to associate a (adjacency) graph to the data
exploiting the neighborhood structure of each datum, given either by
the data points found in small ball centered at the datum itself or by its $k$ nearest
neighbors ($k$NN). An important such method is called isomaps
\cite{TdSL00}. It approximates geodesic distance by minimum path
distance on the graph before using MDS. In this way the global
structure is preserved.  Others are Laplacian Eigenmaps
\cite{BN03} and Diffusion maps \cite{CL06}, where
eigenfunctions (vectors) of the graph Laplacian provide 
an embedding of the data given simply by their evaluation at the given
data point for the first, or weighted evaluation (where the weights are given by
``time'' and the exponential decay of the modes for the corresponding
heat semigroup) for the second. Diffusion maps have the advantage of
uncovering any multiscale structure present, see
\cite{CLLMNWZ05,CKLMN06,LL06} and of providing robustness against
noisy data and/or outliers. Methods based on the graph Laplacian also
admit a solid theoretical foundation as it is known that the graph
Laplacian (sometimes) converges to the Laplace-Beltrami operator of
the manifold that is being sampled \cite{BN08,HAvL07} and that
(carefully selected) eigenfunctions can indeed (rigorously) provide local coordinates, see \cite{JMS10}
for the theoretical result and \cite{KCM21} for an implementation. Interestingly
Laplacian Eigenmaps ideas and techniques have recently been exploited for the
estimation of the tangent space to a manifold as an alternative to a
more direct method based on (local) PCA in a way that is more robust
to noise \cite{KRMC25}.
Here we take a different approach and use an optimization procedure to
obtain a continuous (as opposed to discrete) defining function from
which geometric quantities can be evaluated. The main advantage of
the proposed approach is that it is global in nature, does not require
any structure or organization of the point cloud, and can be deployed
even in the presence of noise. It is often of interest to be able to
deal with new points that were not part of the original data set
\cite{BPVDRO04}. For the proposed method, this is immediate since it
produces a (approximate) defining function for the underlying
manifold that can be evaluated anywhere.
The presented approach borrows ideas from interpolation
theory \cite{We02} and statistical regression \cite{Wah90}. The
connection between these two points of view is discussed in more detail
in \cite{G26}, where the ideas are further developed and lead to (high
order) discretizations of geometric operators on point clouds.
\section{Constructing (Approximate) Defining Functions}
\subsection{Minimal Regularity}
Let $\mathcal{M}\subset \mathbb{R}^d$ be a closed bounded smooth oriented manifold,
which will often be taken to be a (hyper)surface, and denote by
$\delta_\mathcal{M}\in \mathcal{E} '\subset \mathcal{S}'$ the
compactly supported generalized function (tempered distribution, in
fact) defined by 
$$
\langle \delta_\mathcal{M},\varphi\rangle=\int_\mathcal{M}
\varphi(x)\, d \sigma_\mathcal{M} (x),\: \varphi\in
\operatorname{C}^\infty(\mathbb{R}^n),
$$
where $d \sigma_\mathcal{M}$ is the volume form on $\mathcal{M}$. Here
we use the classical notation $\mathcal{E}$ and $\mathcal{S}$ for the
space of smooth ($\operatorname{C}^\infty)$ functions and the space of
rapidly decreasing smooth functions, respectively. The latter consists
of functions $u\in \mathcal{E}$ for which
$$
q_{k,m}(u)=\sup_{|\alpha|\leq k}\sup_{x\in
  \mathbb{R}^d}(1+|x|^2)^{m/2}|\partial^\alpha 
  u(x)|<\infty
$$
for every $m,k\in \mathbb{N}$. Then $\mathcal{E}'$ and $\mathcal{S}'$
are their topological duals as locally convex spaces with seminorms
given by $p_{k,m}=\sup_{|\alpha|\leq k}\sup_{x\in \overline{\mathbb{B}}(0,m)}
  | \partial^\alpha\cdot|$ and $q_{k,m}$, $k,m\in
  \mathbb{N}$. We
are interested in finding a smooth function
$u_\mathcal{M}:\mathbb{R}^d\to \mathbb{R}$ which has the property that
$$
[u_\mathcal{M}=1]\cap \mathcal{M}_\delta =\mathcal{M}\text{ for some
}\delta >0,
$$
where $\mathcal{M}_\delta$ is a neighborhood ($\delta$-fattening) of
$\mathcal{M}$ given by
$$
\mathcal{M}_\delta=[ d(\cdot,\mathcal{M})<\delta].
$$
We take two approaches to generating functions $u_\mathcal{M}$ with
the above property characterized by two extremal choices for their
global regularity. For $s\in \mathbb{R}$, let
$$
\operatorname{H}^s(\mathbb{R}^d)=\big\{ u\in
\mathcal{S}'\,\big |\, (1+|\xi|^2)^{s/2}\hat u\in
\operatorname{L}^2(\mathbb{R}^d) \big\}
$$
be the usual Bessel potential space, where $\hat u$ denotes the
Fourier transform of $u$. First we describe the ``minimal regularity'' case
and consider the (``Laplacian Regularized'') optimization Problems
\begin{equation}\label{lrp0}
\operatorname{argmin}_{u\in
\operatorname{H}^{\frac{d+1}{2}}(\mathbb{R}^d),u|_\mathcal{M}\equiv 1}
\underset{E_0(u) }{\underbrace{\frac{1}{2c_d}\| (1-4\pi^2\Delta)^{\frac{d+1}{4}}u\|_2^2}}\tag{LRP$_0$}
\end{equation}
and, for $\alpha>0$,
\begin{equation}\label{lrpalpha}
\operatorname{argmin}_{u\in
\operatorname{H}^{\frac{d+1}{2}}(\mathbb{R}^d)}\underset{E_\alpha (u) }{\underbrace{\Bigl(
E_0(u)+\frac{1}{2\alpha}\fint_\mathcal{M}
(u-1)^2\, d \sigma_\mathcal{M}\Bigr)}},\tag{LRP$_\alpha$}
\end{equation}
where $c_d=\Gamma(d+1)/\pi^\frac{d+1}{2}$ and
$\fint_\mathcal{M}(\cdot)\, d \sigma _\mathcal{M}
=\frac{1}{|\mathcal{M}|}\int_\mathcal{M} (\cdot)\, d
\sigma_\mathcal{M}$ for $|\mathcal{M}|=\int _\mathcal{M} d
\sigma_\mathcal{M}$. The objective functionals above are denoted by
$E_0$ and by $E_\alpha$ for $\alpha>0$, respectively.
\begin{lem}\label{existence}
The optimization problems \eqref{lrpalpha}, $\alpha\geq 0$, possess a
unique minimizer $u^\alpha
_\mathcal{M}\in\operatorname{H}^{\frac{d+1}{2}}(\mathbb{R}^d)$. For
$\alpha>0$, the minimizer is a weak solution of the equation
\begin{equation}\label{eulera}
A u:=\frac{1}{c_d}(1-4\pi^2\Delta)^{\frac{d+1}{2}}u=
\frac{1}{\alpha|\mathcal{M}|}(1-u)\delta_\mathcal{M},
\end{equation}
i.e. a solution of the equation in $\mathcal{E}'$ (in fact, in
$\operatorname{H}^{-\frac{d+1}{2}}(\mathbb{R}^d))$, or, explicitly it
satisfies
$$
\alpha \langle Au,v \rangle =\frac{\alpha}{c_d} \int
_{\mathbb{R}^d}\bigl[(1-4\pi^2\Delta)^{\frac{d+1}{4}}u\bigr](x)\cdot
  \bigl[(1-4\pi^2\Delta)^{\frac{d+1}{4}}v\bigr](x)\, dx=\fint
  _\mathcal{M} (1-u)(x)v(x)\, d \sigma _\mathcal{M}(x),
$$
for all $v\in \operatorname{H}^{\frac{d+1}{2}}(\mathbb{R}^d)$. If
$\alpha=0$, then it holds that $Au^0_\mathcal{M}$ is a distribution of
order 0 with $\operatorname{supp}(Au_\mathcal{M})\subset \mathcal{M}$,
that is,
\begin{equation}\label{euler0}
Au^0_\mathcal{M}=\psi^0 \frac{1}{|\mathcal{M}|}\delta_\mathcal{M},
\end{equation}
for some $\psi^0\in \operatorname{L}^1(\mathcal{M})$, where
$$
 \langle \psi^0 \delta_\mathcal{M},\varphi \rangle :=\int_\mathcal{M}
 \psi^0(x)\gamma_\mathcal{M}\varphi(x)\, d
 \sigma_\mathcal{M}(x),\:\varphi\in
 \operatorname{H}^{\frac{d+1}{2}}(\mathbb{R}^d),
$$
and $\gamma_\mathcal{M} \varphi$ is the trace of $\varphi$ on
$\mathcal{M}$. 
\end{lem}
\begin{proof}
It is a standard result, \cite{Ada75}, that
$\operatorname{H}^{\frac{d+1}{2}}(\mathbb{R}^d)\hookrightarrow
\operatorname{BUC}^{1/2}(\mathbb{R}^d)$, where the latter space consists of
all bounded and uniformly H\"older continuous functions on $\mathbb{R}^d$
of order $1/2$ endowed with the norm $\|\cdot\|_\infty+[\cdot]_{1/2}$,
where
$$
[u]_{1/2}=\sup_{\tilde x\neq x}\frac{|u(x)-u(\tilde x)|}{|x-\tilde
  x|^{1/2}},\: u\in \operatorname{BUC}^{1/2}(\mathbb{R}^d).
$$
The evaluation of functions in the Bessel space at points in $\mathbb{R}^d$ is
therefore well-defined. Consequently, the constraint
$u\big\vert _\mathcal{M}\equiv 1$ is meaningful. The energy
functional $E_\alpha$, $\alpha\geq 0$ appearing in (LRP$_\alpha$) is
convex and lower semi-continuous regardless of $\alpha\geq 0$. Convex,
lower semi-countinuous 
functionals on a Hilbert space are weakly lower
semi-continuous. Coercivity is also given since bounded subsets
of $\operatorname{H}^{\frac{d+1}{2}}(\mathbb{R}^d)$ are relatively
weakly compact by the Banach-Alaoglu Theorem. These properties ensure
existence, which is unique since the functionals are strictly
convex. For the case $\alpha=0$ it has to be observed that
minimization occurs over the closed convex (and hence weakly closed)
set consisting of functions $u$ for which $u(\mathcal{M})=\{
1\}$. Taking variations in direction of test functions $\varphi\in
\mathcal{D} (\mathbb{R}^d)$ at the minimizer and using the fact
that the operator $(1-4\pi^2\Delta)^{\frac{d+1}{4}}$ is self-adjoint, it is
seen that
\begin{align*}
0&=\left.\frac{d}{dr}\right|_{r=0}
E_\alpha (u^\alpha _\mathcal{M}+r\varphi)=\frac{1}{c_d}\int
(1-4\pi^2\Delta)^{\frac{d+1}{4}}u^\alpha _\mathcal{M}
   (1-4\pi^2\Delta)^{\frac{d+1}{4}}\varphi\, dx+\frac{1}{\alpha}\fint
   _\mathcal{M}(u^\alpha_\mathcal{M}-1)\varphi\, d \sigma_\mathcal{M} \\
  &=\frac{1}{c_d}\big\langle
(1-4\pi^2\Delta)^{\frac{d+1}{4}}u^\alpha_\mathcal{M}
,(1-4\pi^2\Delta)^{\frac{d+1}{4}}\varphi\big\rangle_{\operatorname{L}^2(\mathbb{R}^d)}+
 \frac{1}{\alpha}\big\langle u^\alpha_\mathcal{M}-1,\varphi\big\rangle_{\operatorname{L}^2(\mathcal{M})}\\
  &=\big\langle\frac{1}{c_d}
    (1-4\pi^2\Delta)^{\frac{d+1}{2}}u^\alpha_\mathcal{M},\varphi\big\rangle
  _{\operatorname{H}^{-\frac{d+1}{2}},\operatorname{H}^{\frac{d+1}{2}}}+
  \frac{1}{\alpha}\big\langle u^\alpha_\mathcal{M}-1,\varphi\big\rangle_{\operatorname{L}^2(\mathcal{M})},
\end{align*}
which is valid for all $\alpha\geq 0$ with the understanding that the
manifold term is absent when $\alpha=0$. Taking special test functions
$\varphi\in \mathcal{D}(\mathbb{R}^d\setminus \mathcal{M})$,
i.e. supported away from $\mathcal{M}$, the second term vanishes for
all $\alpha\geq 0$. This shows that
$$
\operatorname{supp}(Au^\alpha_\mathcal{M})\subset\mathcal{M}.
$$
Compactly supported distributions are known to be of finite order. Using that
$$
\operatorname{BUC}^1(\mathbb{R}^d)\varsubsetneq
\operatorname{H}^{\frac{d+1}{2}}(\mathbb{R}^d)\hookrightarrow
\operatorname{BUC}(\mathbb{R}^d),
$$
it is concluded that the order of the distribution $Au^\alpha_\mathcal{M}$
indeed vanishes (the space on the left consists of functions that are
bounded and uniformly continuous, $\operatorname{BUC}$, with all first
partial derivatives enjoying the same property).
When $\alpha>0$, the energy functional contains the
manifold term, so that the above gives
$$
\frac{1}{c_d}\big\langle (1-4\pi^2\Delta)^{\frac{d+1}{4}}u^\alpha_\mathcal{M}
,(1-4\pi^2\Delta)^{\frac{d+1}{4}}\varphi\big\rangle_{\operatorname{L}^2(\mathbb{R}^d)}+
\frac{1}{\alpha|\mathcal{M}|}\big\langle
(u^\alpha_\mathcal{M}-1)\delta_\mathcal{M},\varphi\big
\rangle_{\operatorname{L}^2(\mathcal{M})}=0
$$
for all $\varphi\in \mathcal{D}(\mathbb{R}^d)$ and hence for all
$\varphi\in \operatorname{H}^{\frac{d+1}{2}}(\mathbb{R}^d)$, which
amounts to the claimed equation in weak form.
\end{proof}
Next we exploit the fact that a fundamental solution for the operator
$A$ is known. It is indeed the so-called Laplace kernel $L$ given by
$L(x)=e^{-|x|}$, $x\in \mathbb{R}^d$. This allows us to invert the
operator $A$ by convolution with $L$. In particular, when $\alpha>0$,
this leads to the formulation
\begin{equation}\label{repa}
\alpha u(x)=\fint _\mathcal{M} e^{-|x-y|}(1-u)(y)\, d
\sigma_\mathcal{M}(y),\: x\in \mathbb{R}^d,
\end{equation}
equivalent to \eqref{eulera}\footnote{See the proof of \eqref{rep0}
  for a more detailed explanation of this.}, from which we infer that
the solution $u_\mathcal{M}$ is known if its 
values on $\mathcal{M}$ are known. This makes for a dimensional
reduction akin to that obtained in interpolation via reproducing
kernels where an infinite dimensional problem reduces to a finite
dimensional one (see later discussion of the discrete counterpart of
the current situation). Indeed, evaluating \eqref{repa} at $x\in
\mathcal{M}$, yields the integral equation of the second kind
\begin{equation}\label{iea}
\alpha \gamma_\mathcal{M} u+ \fint _\mathcal{M} L(\cdot-y)
\gamma_\mathcal{M} u(y)\, d
\sigma_\mathcal{M}(y)=\fint _\mathcal{M}L(\cdot-y)\, d
\sigma_\mathcal{M}(y),
\end{equation}
for $\gamma_\mathcal{M} u$ on $\mathcal{M}$. When $\alpha=0$, the
identity corresponding to \eqref{euler0} yields the Ansatz
\begin{equation}\label{laplaceAnsatz}
u^0_\mathcal{M}
=L*\bigl(\psi^0\frac{1}{|\mathcal{M}|}\delta_\mathcal{M}\bigr).
\end{equation}
\begin{lem}\label{representation}
If the density function $\psi^0\in\operatorname{L}^1(\mathcal{M})$ in
\eqref{laplaceAnsatz} is known, then it holds that
\begin{equation}\label{rep0}
 u^0_\mathcal{M}(x) =\fint _\mathcal{M} \psi^0(y)e^{-|x-y|}\, d
 \sigma_\mathcal{M}(y),\: x\in \mathbb{R}^n.
\end{equation}
\end{lem}
\begin{proof}
As observed above, the Laplace kernel $L=e^{-|\cdot|}$ is a fundamental
solution of the (pseudo)differential operator
$\frac{1}{c_d}(1-4\pi^2\Delta)^{\frac{d+1}{2}}$ and thus it holds that
$$
\Bigl(L*(\psi^0\delta_\mathcal{M})\Bigr)(x)=\langle \tau_x\bigl(\psi^0
\delta_\mathcal{M}\bigr), L\rangle=\langle \psi^0
\delta_\mathcal{M}, \tau_x L\rangle.
$$
The convolution is to be understood in the sense of distributions
where $\tau_x$ is translation by $x$, i.e. $\tau_x f=f(\cdot-x)$ for
functions and $\tau_x$ is defined by duality for distributions.
\end{proof}
This lemma yields an equation for the density $\psi^0$ in the form
\begin{equation}\label{ie0}
\fint_\mathcal{M} \psi^0(y)e^{-|x-y|}\, d \sigma_\mathcal{M}(y)=1,\: x\in
\mathcal{M} .
\end{equation}
\begin{rem}
We would like to point an interesting connection out with
research on metric space geometry, where a quantitiy
known as {\em magnitude} (an invariant, when available) plays an
important role. For its computation, one
needs a so-called weighting that parallels the density function
$\psi_0$ in \eqref{ie0} or the
coefficient vector $\Lambda^0$ in \eqref{dies0} below. While in our context, the
appearance of the Laplace kernel is natural in the sense that it is
determined by the choice of regularizer in the optimization problem,
its use in the computation of magnitude is somewhat more
mysterious. It is, however, remarkable that the very same kernel can
encode some geometric information about the underlying metric space.
While it would be of interest to uncover any deeper connections, these
do not appear obvious and we refer the interested reader to the
seminal papers \cite{L08,L13,L19} for the case of finite metric spaces
as well as to
\cite{W09,LW13,W13} for the case of infinite spaces (as the limit of
finite spaces or with the weight defined as a Borel
measure). For so-called positive definite spaces, \cite{M13} showed that
the limiting finite space approach is compatible with the infinite space
one. Notice that subsets of Euclidean space are positive definite in
this sense.
\end{rem}
Equation \eqref{ie0} is a Fredholm integral equation of the first kind. Introducing
a Lagrange multiplier $\Lambda:\mathcal{M}\to \mathbb{R}$ for the
constraint $u\big |_\mathcal{M} \equiv 1$ in \eqref{lrp0} yields the
functional 
$$
E_0(u,\Lambda)=E_0(u)+\frac{1}{|\mathcal{M}|}\langle \Lambda,
1-\gamma_\mathcal{M}u \rangle _\mathcal{M} 
$$
which entails that $u^0_\mathcal{M}$ weakly solves the equation
$$
 Au=\frac{1}{|\mathcal{M}|}\Lambda^0 \delta_\mathcal{M} 
$$
once the Lagrange multiplier $\Lambda^0$ is known. This provides another
justification for the Ansatz and for Equation \eqref{ie0}.
Any solution
yields a critical point of $E_0$, which is necessarily a
minimizer. The equation has a solution since a minimizer
$u^0_\mathcal{M}$ of $E_0$ is a weak solution and it is known
to exist.  The density (or Lagrange multiplier) can be recovered
from $Au^0_\mathcal{M}$ by means of Riesz representation theorem.
As an equation of the first kind, \eqref{ie0} is ill-posed,
while \eqref{iea} is not, as an equation of the second kind. In
fact the latter can be viewed and thought of as a regularization of the
former.
\begin{prop}
Denoting by $u^\alpha _\mathcal{M}$ the minimizer of $E_\alpha$ for
$\alpha\geq 0$, it holds that
$$
u^\alpha_\mathcal{M}  \to u^0_\mathcal{M} \text{ as }\alpha\to 0 \text{ in }
\operatorname{H}^{\frac{d+1}{2}}(\mathbb{R}^d).
$$
\end{prop}
\begin{proof}
Denote the infimum of $E_\alpha$ by $e_\alpha\geq 0$ and notice that
$e_\alpha\leq e_0$ for $\alpha>0$ since $u^0_\mathcal{M}\in
\operatorname{H}^{\frac{d+1}{2}}(\mathbb{R}^d)$ and
$E_\alpha(u^0_\mathcal{M})=e_0$. This entails that
$\| u^\alpha
_\mathcal{M}\|_{\operatorname{H}^{\frac{d+1}{2}}(\mathbb{R}^d)}\leq
C<\infty$ independently of $\alpha>0$. Weak compactness yields a
weakly convergent subsequence with limit $w\in
\operatorname{H}^{\frac{d+1}{2}}(\mathbb{R}^d)$, which, by weak lower
semicontinuity of the norm, must satisfy $\frac{1}{2c_d}\| w\|^2
_{\operatorname{H}^{\frac{d+1}{2}}(\mathbb{R}^d)}\leq e_0$\footnote{We
can work with the norm given by $\| u\|
_{\operatorname{H}^{\frac{d+1}{2}}(\mathbb{R}^d)}=\|
\bigl(1+4\pi^2|\xi|^2\bigr)^{\frac{d+1}{4}}\hat
u\|_{\operatorname{L}^2(\mathbb{R}^d)}$}
and is therefore the unique minimizer $u^0_\mathcal{M}$.
Take now $0<\alpha_0<\alpha_1$ and observe that
\begin{align*}
e_{\alpha_0}=E_{\alpha_0}(u^{\alpha_0}_\mathcal{M})&\leq
[\frac{1}{2\alpha_0}-\frac{1}{2\alpha_1}]
\fint_\mathcal{M}(u^{\alpha_1}_\mathcal{M}-1)^2\, d
\sigma_\mathcal{M}+E_{\alpha_1}(u^{\alpha_1}_\mathcal{M})\\
&\leq [\frac{1}{2\alpha_0}-\frac{1}{2\alpha_1}]
\fint_\mathcal{M}(u^{\alpha_1}_\mathcal{M}-1)^2\, d
\sigma_\mathcal{M}+E_{\alpha_1}(u^{\alpha_0}_\mathcal{M})
\end{align*}
implies that $\fint_\mathcal{M}(u^{\alpha_0}_\mathcal{M}-1)^2\, d
\sigma_\mathcal{M}\leq \fint_\mathcal{M}(u^{\alpha_0}_\mathcal{M}-1)^2\, d
\sigma_\mathcal{M}$. It follows that
\begin{align*}
\|
u^{\alpha_1}_\mathcal{M}\|_{\operatorname{H}^{\frac{d+1}{2}}(\mathbb{R}^d)}&\leq
\|u^{\alpha_0}_\mathcal{M}\|_{\operatorname{H}^{\frac{d+1}{2}}(\mathbb{R}^d)}
-\frac{1}{2 \alpha_1}\bigl[ \fint_\mathcal{M}(u^{\alpha_1}_\mathcal{M}-1)^2\, d
\sigma_\mathcal{M}-\fint_\mathcal{M}(u^{\alpha_0}_\mathcal{M}-1)^2\, d
\sigma_\mathcal{M}\bigr]\\&\leq\|
  u^{\alpha_0}_\mathcal{M}\|_{\operatorname{H}^{\frac{d+1}{2}}(\mathbb{R}^d)}.
\end{align*}
Finally take any null sequence $(\alpha_n)_{n\in\mathbb{N}}$. By the
above weak compactness and lower semicontinuity argument, it will have
a subsequence that converges to $u^0_\mathcal{M}$. Since,
additionally, the norm converges thanks to
$$
\|u^0_\mathcal{M}\|_{\operatorname{H}^{\frac{d+1}{2}}(\mathbb{R}^d)}\leq
\liminf_{k\to\infty}\|
u^{\alpha_k}_\mathcal{M}\|_{\operatorname{H}^{\frac{d+1}{2}}(\mathbb{R}^d)}
\leq \limsup_{k\to\infty}\|
u^{\alpha_k}_\mathcal{M}\|_{\operatorname{H}^{\frac{d+1}{2}}(\mathbb{R}^d)}
\leq \| u^0_\mathcal{M}\|_{\operatorname{H}^{\frac{d+1}{2}}(\mathbb{R}^d)},
$$
the subsequence actually converges in norm. Since this is true for any
null sequence $(\alpha_n)_{n\in\mathbb{N}}$ (with the same limit
$u^0_\mathcal{M}$), the whole family
converges in norm as claimed.
\end{proof}
This natural regularization will prove very useful in numerical
applications.
\begin{prop}\label{ansatzProp}
Using the same Ansatz
\begin{equation}\label{ansatza}
u^\alpha_\mathcal{M} =\fint _\mathcal{M}
\Lambda^\alpha(y)e^{-|\cdot-y|}\, d \sigma_\mathcal{M}(y)
\end{equation}
when $\alpha>0$, it holds that
$$
u^\alpha _\mathcal{M} \big |_\mathcal{M} =\fint_\mathcal{M} 
\Lambda^\alpha(y)e^{-|\cdot-y|}\, d \sigma_\mathcal{M}(y)
=1-\alpha \Lambda^\alpha,
$$
i.e. the validity of an equation for the density $\Lambda^\alpha$ given by
\begin{equation}\label{numiea}
 (\alpha +\mathcal{L})\Lambda = 1\text{, where }\mathcal{L} \Lambda
 :=\fint _\mathcal{M} e^{-|\cdot -y|}\Lambda(y)\, d \sigma_\mathcal{M}(y)
\end{equation}
\end{prop}
\begin{proof}
It follows from the Ansatz and from \eqref{eulera} that
$$
\frac{1}{|\mathcal{M}|}\alpha \Lambda \delta_\mathcal{M}=
\frac{1}{|\mathcal{M}|}\bigl(1-u^\alpha_\mathcal{M}\big
|_{\mathcal{M}}\bigr)\delta_\mathcal{M},
$$
which yields an equation for the densities amounting to
\eqref{numiea}.
\end{proof}
\begin{rem}
Representation \eqref{ansatza} and equation \eqref{numiea} are the
most convenient for use in numerical calculations (and will be used in
the numerical experiments presented later).
\end{rem}
\begin{rem}
The representations \eqref{repa} and \eqref{rep0} show that
$$
u^\alpha_\mathcal{M} \in \operatorname{C}^\infty(\mathbb{R}^d\setminus
\mathcal{M} )\cap \operatorname{H}^{\frac{d+1}{2}}(\mathbb{R}^d),
$$
and therefore has generically smooth level sets away from
$\mathcal{M}$. While $\mathcal{M}$ is itself a level set, its
regularity is clearly what it is. Notice that, while $u^\alpha_\mathcal{M} $
is smooth away from $\mathcal{M}$, its low global regularity allows
for a sharp transition in values moving away from $\mathcal{M} $.
\end{rem}
\subsection{High Regularity}
Inspired by the variational problems (LRP$_0$) we try to find a
high regularity $u^0 _\mathcal{M} \in
\operatorname{C}^\infty(\mathbb{R}^d)$ by considering the problem
\begin{equation}\label{heatEq}
  \begin{cases} u_t-\Delta u=0,&\text{in }(0,\infty)\times \mathbb{R}^d,\\
  u(0,\cdot)=\frac{\pi^\frac{d}{2}}{|\mathcal{M}|}\psi^0 \delta_\mathcal{M},&\text{in
  }\mathbb{R}^d.
\end{cases}
\end{equation}
Its solution $u\in \operatorname{C}^\infty\bigl( (0,\infty)\times
\mathbb{R}^d\bigr)$ is analytic and given by
$$
u(t,x)=\frac{1}{(4t)^\frac{d}{2}}\fint_{\mathcal{M}}e^{-|x-y|^2/4t}\psi^0(y)\, d
\sigma_\mathcal{M}(y),\: (t,x)\in(0,\infty)\times \mathbb{R}^d.
$$
as follows from knowledge of the heat kernel. We obtain a defining
function in the form $u^0_\mathcal{M}=u(\frac{1}{4},\cdot)$ for
$\mathcal{M}$ by imposing the condition 
$$
u(\frac{1}{4},x)=1,\: x\in \mathcal{M},
$$
which, again, amounts to an integral equation of the first kind and
reads
\begin{equation}\label{rie0}
\fint _{\mathcal{M}} e^{-|x-y|^2}\psi^0(y)\, d
\sigma_\mathcal{M}(y)=1,\: x\in \mathcal{M} .
\end{equation}
Changing the initial datum to
$\frac{\pi^{\frac{d}{2}}}{\alpha|\mathcal{M}|}\bigl[
1-u(\frac{1}{4},\cdot)\bigr]\delta_\mathcal{M}$ for $\alpha>0$, which
``penalizes'' deviation from the  value 1, leads the corresponding
regularized problem
\begin{equation}\label{riea}
\alpha u^\alpha_\mathcal{M}(x) +\fint _{\mathcal{M}}
e^{-|x-y|^2}u^\alpha_\mathcal{M} (y)\, d
\sigma_\mathcal{M}(y)=\fint _{\mathcal{M}} e^{-|x-y|^2}\, d
\sigma_\mathcal{M}(y),\: x\in \mathcal{M} .
\end{equation}
Similar to the case of the Laplace kernel, for the Gauss kernel, we
can make the Ansatz
$$
u^\alpha_\mathcal{M}(x)=\fint _{\mathcal{M}}\Lambda ^\alpha(y) e^{-|x-y|^2}\, d
\sigma_\mathcal{M}(y),\: x\in \mathbb{R}^d,
$$
derive the regularized equation
\begin{equation}\label{numriea}
(\alpha+\mathcal{G})\Lambda^\alpha
=1\text{, where }\mathcal{G} \Lambda^\alpha =\fint
_\mathcal{M} e^{-|\cdot-y|^2}\Lambda^\alpha (y)\, dy,
\end{equation}
and observe that $u^\alpha_\mathcal{M}=1-\alpha \Lambda^\alpha$ on
$\mathcal{M}$.
In both frameworks, the minimal regularity and the smooth ones, one
obtains a defining function when $\alpha=0$ and an approximate
defining function when $\alpha>0$. As we shall demonstrate later, even
an approximate defining function can play an important role in
search of geometric information from data sets.
\begin{rem}
A formal but maybe more transparent way to interpret the smooth
approach just described is to introduce the energy functional
$$
G_\alpha(u)=\frac{1}{2}\|e^{-\frac{1}{8}\Delta}u\|^2_{\operatorname{L}^2(\mathbb{R}^d)}
+\frac{\pi^{d/2}}{2\alpha}\fint _\mathcal{M} (u-1)^2\, d \sigma_\mathcal{M},
$$
where the manifold term is replaced by the constraint $u|_\mathcal{M}
\equiv 1$ when $\alpha=0$. Formally its Euler equation is
$$
e^{-\frac{1}{4}\Delta}u=\frac{\pi^{d/2}}{\alpha|\mathcal{M}|}(1-u)\delta_\mathcal{M} 
$$
\end{rem}
\begin{rem}
While the approach described so far is easier to understand
mathematically for  closed compact orientable hypersurfaces $\mathcal{M}$, it
only requires the set $\mathcal{M}$ to admit integration over
it. Numerical experiments will be shown later.
\end{rem}
\subsection{The Kernels and the Equations}
An important property of the Laplace and the Gauss kernels is their
positive definitess in the sense that the matrix
$$
\bigl[ K(x^i-x^j)\bigr]_{1\leq i,j\leq m},\: K=L,G,
$$
is positive definite for any choice of $m$ distinct points
$x^1,\dots,x^m$ in $\mathbb{R}^d$ and any choice of $m\in
\mathbb{N}$. This follows from Bochner's Theorem 
as the Fourier transforms $\widehat L$ and $\widehat G$ of these kernels
are positive and integrable functions. At the continuous level, given
a compact orientable smooth manifold $\mathcal{M}\subset \mathbb{R}^d$,
the right-hand sides of \eqref{ie0} and of \eqref{rie0} define an
integral operator $\mathcal{K}$ with kernel $K=L,G$
$$
 \mathcal{K}\psi=\int_\mathcal{M} K(\cdot-y)\psi(y)\, d \sigma_\mathcal{M}(y),
 \:\psi \in \operatorname{C}(\mathbb{R}^d)``\subset
 \operatorname{L}^2(\mathcal{M})\text{''}. 
$$
Noticing that
$$
\langle \psi, \mathcal{K}\psi \rangle_{\operatorname{L}^2(\mathcal{M})}=\langle
\psi \delta _\mathcal{M},
K*(\psi\delta_\mathcal{M})\rangle_{\mathcal{E}'_0,\mathcal{E}_0},
$$
where the duality pairing on the right-hand side is that between
compactly supported measures (zero order distributions) and continuous
functions. Using (a generalization of) Plancherel's Theorem in inner product form, we see that
$$
\langle \psi, \mathcal{K}\psi\rangle_{\operatorname{L}^2(\mathcal{M})}=
\langle\widehat{\psi \delta _\mathcal{M}},\widehat{K}\,\widehat{\psi \delta
  _\mathcal{M}}\rangle_{\operatorname{L}^2(\mathbb{R}^d)}>0,
$$
by the Paley-Wiener Theorem for compactly supported distributions and
by the positivity and integrability (decay properties) of the Fourier transform of the
kernel $K$. Thus there cannot exist a nontrivial function $\psi$ in
the nullspace of $\mathcal{K}$ and equations \eqref{ie0} and
\eqref{rie0} are uniquely solvable. As $\mathcal{K}$ is a compact
operator, the inverse is clearly unbounded and the problems
ill-posed. In this sense \eqref{iea} and \eqref{riea} can be thought
of as regularizations of \eqref{ie0} and \eqref{rie0}, respectively.
\section{The Discrete Counterpart: Point Clouds}
We now turn our attention to sets of points (point clouds) that are
assumed to be an exact or corrupted sample of the points of a manifold
$\mathcal{M}\subset \mathbb{R}^d$, where, in the later experiments,
$d=2,3$. Let $\mathbb{X}\subset \mathbb{R}^d$ denote a finite subset
of disinct points (possibly sampled from an underlying manifold $\mathcal{M}$)
which we think as listed, i.e. $\mathbb{X}=\big\{ x^i\,\big |\,
i=1,\dots,m\big\}$, but where the order has no particular meaning. If
this is all the information we are given about the manifold
$\mathcal{M}$, it is natural to use these points as collocation
points to approximate the various continuous quantities and equations
of the previous section. In particular, it is natural to approximate
the normalized measure
\begin{equation}\label{dmeasure}
\frac{1}{|\mathcal{M}|}\delta_\mathcal{M} \simeq
\frac{1}{m}\sum_{i=1}^m \delta_{x^i},
\end{equation}
by the empirical measure obtained from $\mathbb{X}$. Then equations
\eqref{ie0} and \eqref{rie0} can be approximated by
\begin{equation}\label{dies0}
\frac{1}{m}\sum_{k=1}^m K(x^i-x^k)\Lambda^0(x^k)=1,\: i=1,\dots,m.
\end{equation}
and equations \eqref{numiea} and \eqref{numriea} by
\begin{equation}\label{driesa}
\alpha \Lambda^\alpha_m(x^i)+\frac{1}{m}\sum_{k=1}^m
K(x^i-x^k)\Lambda^\alpha(x^k)=1,\: i=1,\dots,m,
\end{equation}
where $K$ is the Laplace kernel $L$ and Gauss kernel $G$,
respectively. Both these kernels are positive definite and thus the
matrix
\begin{equation}\label{matrix}
\mathbb{R}^{m\times m}\ni M=\bigl[ K(x^i-x^k)\bigr]_{1\leq i,k\leq m}
\end{equation}
is positive definite and hence invertible. Denoting the discrete
density by $\Lambda^\alpha_m\in \mathbb{R}^m$ (which can be thought of
as an approximation for $\Lambda^\alpha(x^k)$, $k=1,\dots,m$) we obtain
the (approximate when $\alpha>0$) signature (defining) function
of the point cloud $\mathbb{X}$ by 
\begin{equation}\label{numdefFct}
u_\mathbb{X}(x)=\frac{1}{m}\sum_{k=1}^m \Lambda^{\alpha,k}_mK(x-x^k),\: x\in
\mathbb{R}^d,
\end{equation}
which has the important advantage of being defined and
numerically computable everywhere.
\begin{prop}
The signature function $u_\mathbb{X}$ depends continuously on the data
set $\mathbb{X}$ with respect to the
$\operatorname{H}^r(\mathbb{R}^d)$ norm for $r=\frac{d+1}{2}$, when
using the Laplace kernel and any $r\in \mathbb{N}$, when using the Gauss
kernel. More precisely, the map
$$
(x^1,\dots,x^m)\mapsto u_\mathbb{X},\: \mathbb{R}^m\to
\operatorname{H}^r(\mathbb{R}^d),
$$
where $\mathbb{X}=\{ x^1,\dots, x^m\}$, is continuous.
\end{prop}
\begin{proof}
We consider the case of the Laplace kernel first. The claim follows
from the fact that, once the values of $u_\mathbb{X}$ on 
$\mathbb{X}$ are established, equations \eqref{eulera}-\eqref{euler0}
can be used together with the continuous dependence of their solutions
on the right hand-hand side. The latter depends itself continuously on
the data set $\mathbb{X}$ because the Dirac distribution $\delta_x$
depends continuously on its location $x\in \mathbb{R}^d$ in the
topology of $\operatorname{H}^{-\frac{d+1}{2}}(\mathbb{R}^d)$ as
follows from the Sobolev embedding
$$
\| \delta_x-\delta_{\tilde x}\|_{\operatorname{H}^{-\frac{d+1}{2}}}=
\sup_{\|\varphi\|_{\operatorname{H}^{\frac{d+1}{2}}}=1}\big |\langle \delta_x-
\delta_{\tilde x} , \varphi\rangle\big |=
\sup_{\|\varphi\|_{\operatorname{H}^{\frac{d+1}{2}}}=1}\big |\varphi(x)-\varphi(\tilde x)\big |\leq
C\,|x-\tilde x|^{\frac{1}{2}},\: x,\tilde
x\in \mathbb{R}^d
$$
thanks to $\sup_{x\neq
  \tilde x}\frac{|\varphi(x)-\varphi(\tilde x)|}{|x-\tilde x|^{1/2}}\leq C\| \varphi\|_{\operatorname{H}^{\frac{d+1}{2}}}$
and because the values of $u_\mathbb{X}$ on $\mathbb{X}$ are obtained from
a linear system, the matrix of which depends continuously on
$\mathbb{X}$ as well. A similar reasoning can be applied when the
chosen kernel is Gaussian. In that case, we use the continuous
dependence of the solution of the heat equation on its initial data in
the weak norm and the regularizing effect of the heat equation for
positive times. The initial data are again a finite linear combination of
Dirac distributions supported on the data set, the coefficients of
which also depend continuously on the data set.
\end{proof}
\begin{rem}
Thinking of equations \eqref{dies0}\&\eqref{driesa} as discretizations of
equations \eqref{ie0}\&\eqref{rie0} and equations
\eqref{numiea}\&\eqref{numriea} (depending on the choice of kernel), it
would appear, that that weight of the surface measure $d
\sigma_\mathcal{M}$ has been neglected (and it 
has), but since the focus is on the signature function $u_\mathbb{X}$,
there is no need to have direct access to it and we can think of it
as being incorporated in the density function $\Lambda^\alpha_m$. 
\end{rem}
\begin{rem}
Interpreting equations \eqref{dies0}\&\eqref{driesa} as discretization
of their continuous counterpart is useful for the structural
understanding of the problem. It is, however, remarkable that the
solutions $u_\mathbb{X}$ produced by equations
\eqref{dies0}\&\eqref{driesa} are themselves minimizers of an infinite
dimensional optimization problem. Take the Laplace kernel case, for
instance. Then $u_\mathbb{X}$ is the minimizer of the energy
$$
E_{\alpha,\mathbb{X}}(u)=E_0(u)+\frac{1}{2\alpha m}\sum_{i=1}^m\big |
u(x^i)-1\big |^2,
$$
with the understanding that the second term is replaced by the
constraint $u |_\mathbb{X} \equiv 1$ when $\alpha=0$. This
optimization problem  is well-defined for any data set $\mathbb{X}$ since 
$\operatorname{H}^{\frac{d+1}{2}}(\mathbb{R}^d)\hookrightarrow
\operatorname{BUC}(\mathbb{R}^d)$. It was computed in \cite{G24} for
$\alpha>0$ that the corresponding Euler-Lagrange equation is given by
$$
\alpha Au= \frac{1}{m}\sum_{i=1}^m \bigl[ 1-u(x^i)\bigr] \delta_{x^i}.
$$
This connection also explains the efficacy of the use of the level
sets of $u_\mathbb{X}$ for classification purposes demonstrated in
\cite{G24}. A similar discussion applies in the case of the Gauss
kernel. Indeed, the function
$$
\frac{1}{m}\sum_{k=1}^me^{-|x-x_k|^2}\Lambda^k_m
$$
is the solution of the heat equation evaluated at time $t=\frac{1}{4}$
with initial datum
$$
\frac{\pi^{d/2}}{m}\sum_{k=1}^m\Lambda^k_m \delta_{x^k},
$$
thus yielding a direct intepretation of equation \eqref{dies0} (and
similarly of equation \eqref{driesa}) when $K=G$.
\end{rem}
\begin{rem}
As a matter of fact, the discrete case can be subsumed to the general
case by simply setting $\mathcal{M}=\mathbb{X}$ and setting
$$
\delta_\mathcal{M}=\delta_\mathbb{X}:=\frac{1}{m}\sum_{i=1}^m \delta_{x^i}.
$$
When $\mathbb{X}$ is used or thought of as an approximation of a
continuous manifold $\mathcal{M}$, however, this interpretation is not
always fully compatible with convergence as the discrete points ``fill'' the
manifold $\mathcal{M}$ as explained above.
\end{rem}
\begin{rem}
We point out that equations like \eqref{dies0} and \eqref{driesa} written as
$$
M \Lambda = m{\bf 1}_m\text{ and }(m\alpha+M)\Lambda =m{\bf 1}_m
$$
respectively, arise and have been extensively studied and used for
interpolation and statistical purposes for general right-hand side
(above ${\bf 1}_m$ denotes the vector of length $m$ with all
components 1) . The first as the natural equation for the
computation of an interpolant via the method of Reproducing Kernel
Hilbert Spaces \cite{W04}, and the second as a Ridge Regression in
statistics \cite{Wah90}. In the context of point clouds \cite{BTSAGSS17} and the
implicit representation of hypersurfaces, it appears customary
to find local neighborhoods and use methods like principal component
analysis to obtain an approximate tangent plane (and normal vector) in
order to construct a defining function by prescribing its values at
off-surface points. This approach, often based on the identification
of $k$ nearest neighbors and principal component analysis, is 
described e.g. in \cite{W04} as an application of meshless interpolation
methods. The framework developed here, shows how kernel methods of the
kind discussed in \cite{W04} can, in fact, give direct access to the
geometry of the surface simply using a (not necessarily ordered)
sample of the points on (or near) it. For a more thorough discussion
of these connections we refer to \cite{G26}.
\end{rem}
\begin{rem}
While we will not pursue this angle in this paper further, we point
out that the proposed approach, while global in nature, can be
modulated to possess varying degrees of non-locality. The kernel $K$
can indeed be replaced by $K(\delta\cdot)$ for $\delta>0$, which determines the
width of its bump.
\end{rem}
\section{Symmetries}
In this short section we mainly remark that the signature function
$u_\mathbb{\mathcal{M}}$ inherits any symmetries enjoyed by the manifold
$\mathcal{M}$.
\begin{prop}
Let $R:\mathbb{R}^d\to \mathbb{R}^d$ be any rigid transformation with
the property that $R(\mathcal{M})=\mathcal{M}$. Then
it holds that
$$
u_\mathcal{M}(R\,\cdot)=u_\mathcal{M}. 
$$
\end{prop}
\begin{proof}
For the Laplace case, notice that the energy functional $E_0$
satisfies $E_0(u)=E_0(u\circ R)$ so that $u$ and $u\circ R$ are both
minimizers since they both satisfy the constraint (which is also
invariant under any self-map of $\mathcal{M}$). Uniqueness then implies
that they coincide. When $\alpha>0$, the additional term in the
functional satisfies 
$$
\fint_\mathcal{M}f(Rx)\, d \sigma_\mathcal{M} (x)=\fint
_{R^{-1}\mathcal{M}}f(Rx)\, d \sigma_\mathcal{M}(Rx)= \fint
_\mathcal{M} f(x) \, d \sigma_\mathcal{M}(x,)
$$
for any $f\in \operatorname{L}^1(\mathcal{M})$ since
$\mathcal{M}=R^{-1}\mathcal{M}$ and
$R_*\sigma_\mathcal{M}=\sigma_\mathcal{M}$ by assumption, where here
and below $R^*$ and $R_*$ are the pull-back and push-forward by $R$,
respectively. For the Gaussian case, notice that equation
\eqref{numriea} is invariant with respect to $R$
$$
{\bf 1}_\mathcal{M}=R_*{\bf 1}_\mathcal{M}=
R_*\bigl((\alpha+\mathcal{G})\Lambda\bigr)=R_*(\alpha+\mathcal{G})R^*R_*\Lambda
=(\alpha+\mathcal{G})R_*\Lambda,
$$
since
$$
\fint_\mathcal{M} e^{-|Rx-z|^2}f(R^{-1}z)\, d
\sigma_\mathcal{M}(z)=\fint_{R^{-1}\mathcal{M}}e^{-|Rx-Ry|^2}f(y)\, d
\sigma_\mathcal{M}(Ry)=\fint_\mathcal{M}e^{-|x-y|^2}f(y)\, d
\sigma_\mathcal{M}(y).
$$
Uniqueness again yields the claim together with the fact that the
restriction of the signature function to $\mathcal{M}$ is directly
related to the density $\Lambda^\alpha$ in the Gaussian case via
$$
u^\alpha_\mathcal{M} \big |_{\mathcal{M}}=1-\alpha \Lambda^\alpha,
$$
just as in the Laplace case, a fact which was noted in Proposition \ref{ansatzProp}.
When the data set $\mathbb{X}$ enjoys a symmetry, it will therefore be
reflected in its signature function $u_\mathbb{X}$.
\end{proof}
When the data set is a sample of a continuous manifold, the symmetry
properties of the manifold will be approximately reflected in the
signature function of the sample, as well. This is apparent in some of
the experiments considered in the last section. As already mentioned
in passing above, in some cases the signature function is not a
regular defining function in the sense that the value 1 is not a
regular value for it. This happens when the manifold is flat, for
instance. A simple example is given by the real line
$\mathbb{R}\times\{0\}\subset\mathbb{R}^d$ for which, taking $\alpha=0$
and using the Gaussian kernel we see that
$$
\int _{-\infty}^\infty e^{-|x_1-y_1|^2-|x'|^2}\Lambda(y_1)\,
dy_1=e^{-|x'|^2}\int _{-\infty}^\infty e^{-|x_1-y_1|^2}\Lambda(y_1)\, dy_1,
$$
where $x'=(x_2,\dots,x_d)$. We know that the signature function and
hence the density $\Lambda$ is translation invariant and hence
constant on the real line so that
$$
\Lambda = \bigl( \int _{-\infty}^\infty e^{-|x_1-y_1|^2}\,dy_1\bigr)^{-1}
=\bigl( \int _{-\infty}^\infty e^{-y_1^2}\,dy_1\bigr) ^{-1}.
$$
Then 1 is not a regular value of $u_\mathcal{M}=e^{-|x'|^2}$. It is,
however, possible to compute normals by moving even only slightly
away from the line at any point of interest along it. In general, the
presence of curvature will ensure that this does not happen as values
on one side of a hypersurface will be higher than on the other as
would be the case in the example, if the line were bent and $d=2$. The
example of the line also shows how the signature function of a
manifold falls off rapidly and generates rapidly turning normals in
any direction that is not tangent to it. This will be expolited in
oder to estimate the local dimension of a manifold from its samples. 

\section{The Geometry of Point Clouds}
In this section we use the signature functions obtained in the previous
sections to analyze the normal and the curvature of manifolds
sampled at finitely many points. If the starting point is a
point cloud, then its signature function yields an interpolated
continuous manifold of which geometric quantities can be
computed and used to understand the point cloud itself. The idea is
straightforward: if the point cloud is known or for some reason
supposed to be ``smooth'', then the use of the Gauss kernel is most
appropriate, while, in cases where the surface is known to possess
only low regularity, the best choice is the Laplace kernel. This
point will be further discussed in the next section, where the noisy
situation is considered as well and regularization plays an even
more important role. We discuss the approach for the case of hypersurfaces
and for the Gauss kernel first because of its higher degree of
smoothness. Given a point cloud $\mathbb{X}$ of size $m$, we compute
the associate density function $\Lambda^\alpha_m\in \mathbb{R}^m$ by
solving equation \eqref{driesa} and obtain
$$
u_\mathbb{X}(x)=\frac{1}{m}\sum_{j=1}^m
\lambda_je^{-|x-x^j|^2},\: x\in  \mathbb{R}^d,
$$
where, from now on,  we set $\lambda_j=\Lambda^{\alpha,j}_m$ for
simplicity of notation. Then we compute the normal $\nu_\mathbb{X}$ as
$$
\nu_\mathbb{X}=-\frac{\nabla u_\mathbb{X}}{|\nabla u_\mathbb{X}|},
$$
which is of particular interest at $x\in \mathbb{X}$, where
$$
\nabla u_\mathbb{X}(x)=-2\sum_{j=1}^m \lambda_j e^{-|x-x^j|^2}(x-x^j),
x\in \mathbb{R}^d
$$
and where the sign is chosen in order to obtain the outer unit normal
in the case of a unit circle. Next we compute the  Hessian of
$u_\mathbb{X}$
$$
\bigl[ D^2u_\mathbb{X}\bigr]_{kl}=4\sum_{j=1}^m\lambda_je^{-|x-x^j|^2}
(x_k-x_k^j)(x_l-x_l^j)- 2\delta_{kl}\sum_{j=1}^m\lambda_je^{-|x-x^j|^2},
$$
from which a direct calculation yields the Jacobian of $\nu_{\mathbb{X}}$
$$
D\nu_\mathbb{X} = \frac{1}{|\nabla u_\mathbb{X}|}\bigl( 
D^2u_\mathbb{X}-D^2u_\mathbb{X}\nu_\mathbb{X}\nu_\mathbb{X}^\mathsf{T}\bigr).
$$
Now $\nu_\mathbb{X}$ is clearly an eigenvector of $D\nu_\mathbb{X}$ to
the eigenvalue 0, while the other eigenvalues are the principal
curvatures of the surface. Points at which $u_\mathbb{X}$ is not a regular
defining function, i.e. at which $\nabla u_\mathbb{X}$ vanishes,
cannot be excluded but a small perturbation can be applied numerically
to still obtain a meaningful approximation and avoid the singularity
as discussed previously. When the point cloud represents a full measure
or open subset of the ambient space, the signature function is
better thought of a smooth approximation of its characteristic
function. When the codimension is higher than one, the method still
computes a normal to the interpolated manifold and its curvatures. The
latter are, however, found along with spurious curvatures due to the
fact that the method always generates hypersurfaces for most of its
level sets. In the case of the real line in higher dimensional space,
it follows from the example at the end of the previous section that
most level sets are cylinders.

The Laplace kernel would appear not to be a viable option for the
computation of normals and curvatures due to its lack of
smoothness. In pratice, however, it is sufficient to use a slight
regularization of the kernel given by
$$
L_r(x)=e^{-\sqrt{|x|^2+r}},\: x\in \mathbb{R}^d,
$$
for $r>0$ small, which leads to the well-defined expressions
$$
\partial_j L_r(x)=-e^{-\sqrt{|x|^2+r}}\frac{x^j}{\sqrt{|x|^2+r}},x\in
\mathbb{R}^d,\: j=1,\dots, d
$$
and
$$
\partial_i\partial_jL_r(x)=e^{-\sqrt{|x|^2+r}}\Bigl(
\frac{x^ix^j}{|x|^2+r} +\frac{\delta_{ij}}{\sqrt{|x|^2+r}}
-\frac{x^ix^j}{(|x|^2+r)^{3/2}}\Bigr),\: x\in \mathbb{R}^d,\: i,j=1,\dots, d,
$$
that can be used as in the calculations described above for the case
of the Gauss kernel. This approach consistently yields good estimates
for the normal and, when the data set is dense enough (denser than the
scale determined by $r>0$), also computes viable curvature approximations.
\begin{rem}
It is important to stress the fact that this approach does not require
any organization of the points in the point cloud. The use of the
kernel implicitly takes advantage of local neighborhoods while
maintaing a global significance by including the influence of every
single point in the cloud. 
\end{rem}
\begin{rem}
The simplest point cloud consists of a single point $x_0\in
\mathbb{R}^d$, in which case $u_\mathbb{X}$ yields a function peaked
at $x_0$ and possessing spheres centered at $x_0$ as its level sets
(regardless of the choice of kernel). In the general case, these
building blocks combine in a way determined by the geometry of the
point cloud and the underlying PDE to yield a meaningful signature function.
\end{rem}
\begin{rem}
The regularization approach described above for the Laplace kernel
will be followed in the numerical implementation of the experiments
performed in the next section.
\end{rem}
\section{Numerical Experiments}
In this section we demonstrate how the two ``extremal'' methods
described in the previous sections can be successfully used for the
purposes described and highlight their distinct specific behaviors and
advantages.
\subsection{A Circle} 
\begin{figure}
  \includegraphics[scale=.45]{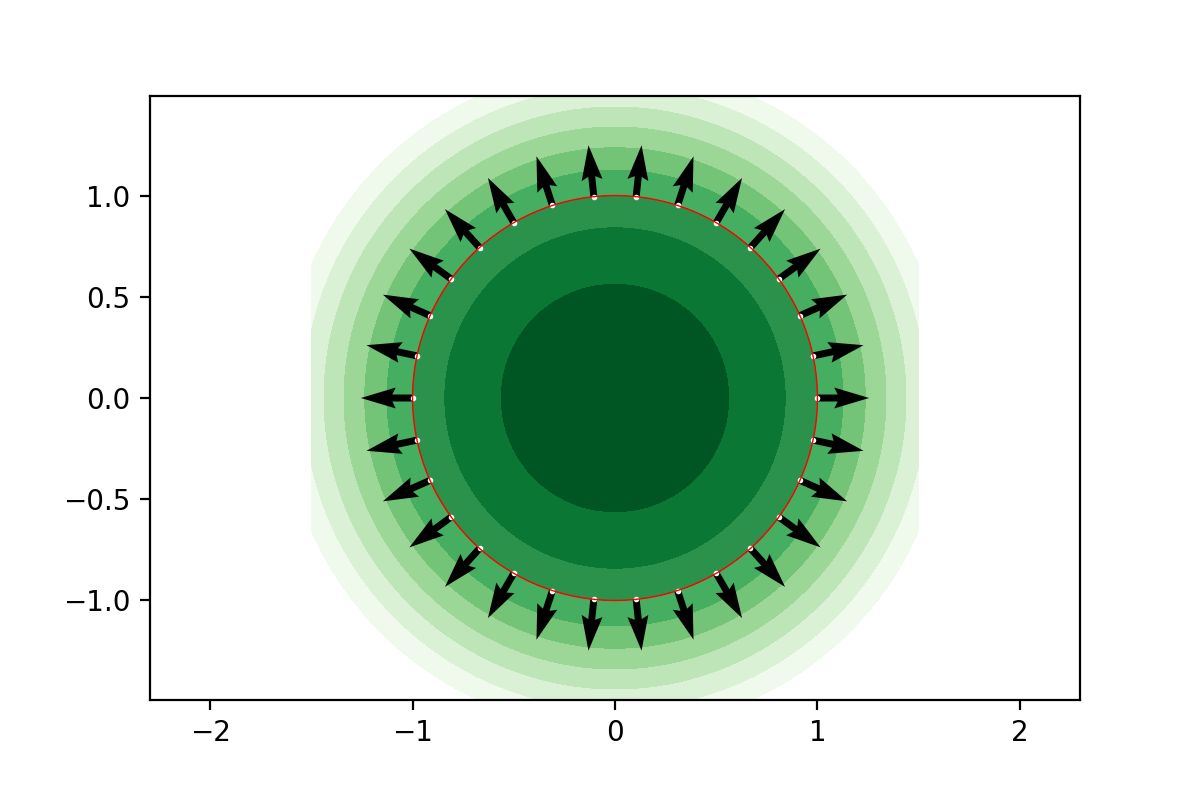}
  \includegraphics[scale=.45]{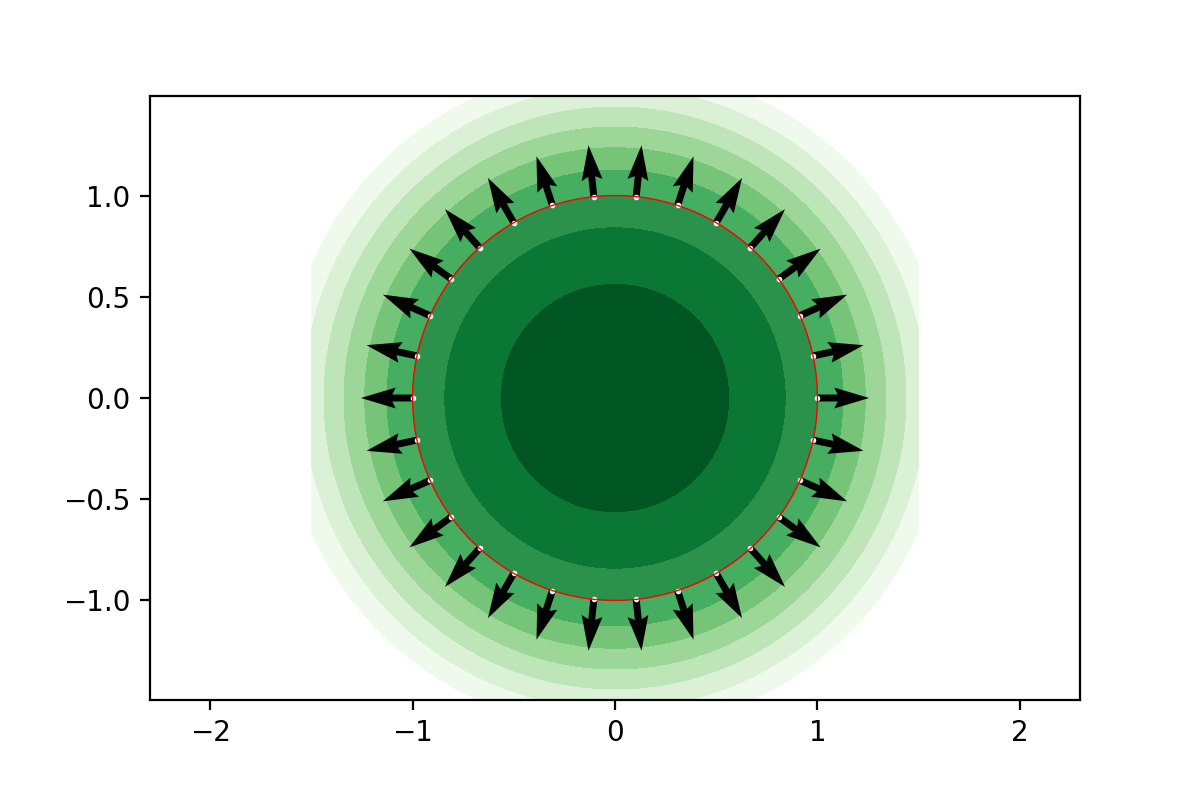}
  \includegraphics[scale=.45]{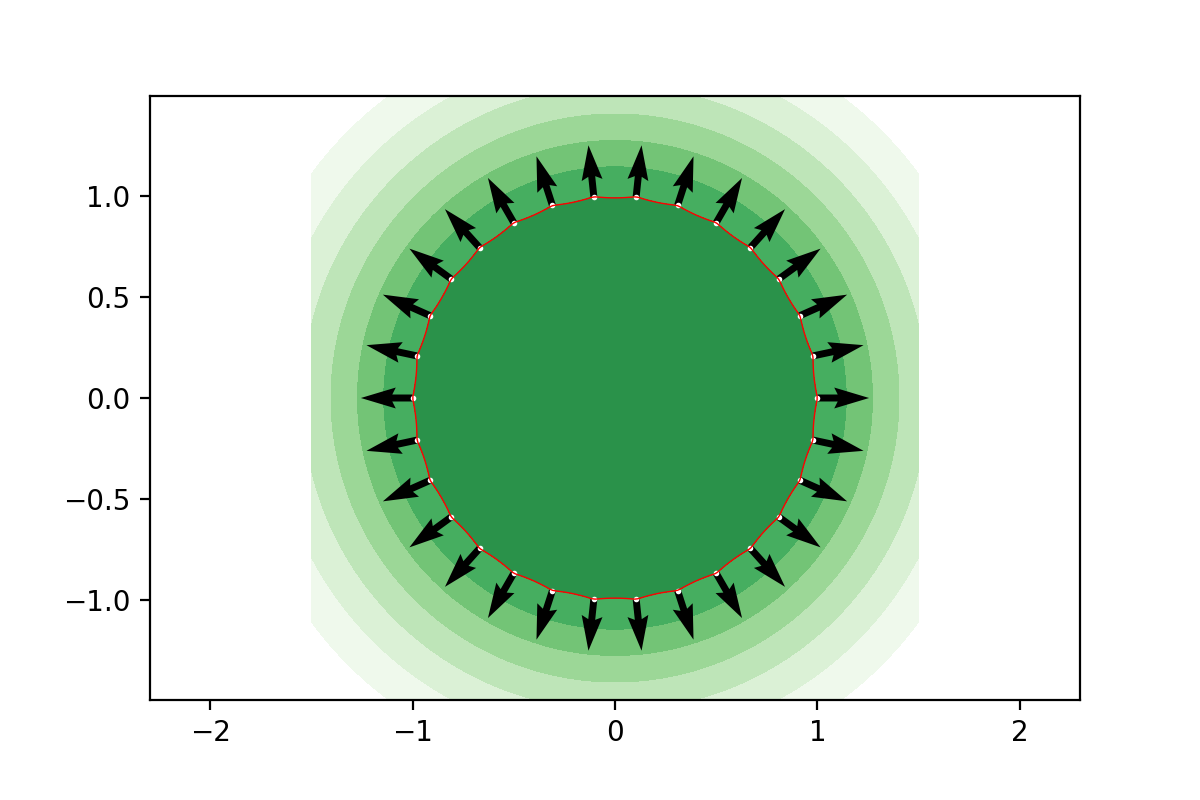}
  \includegraphics[scale=.45]{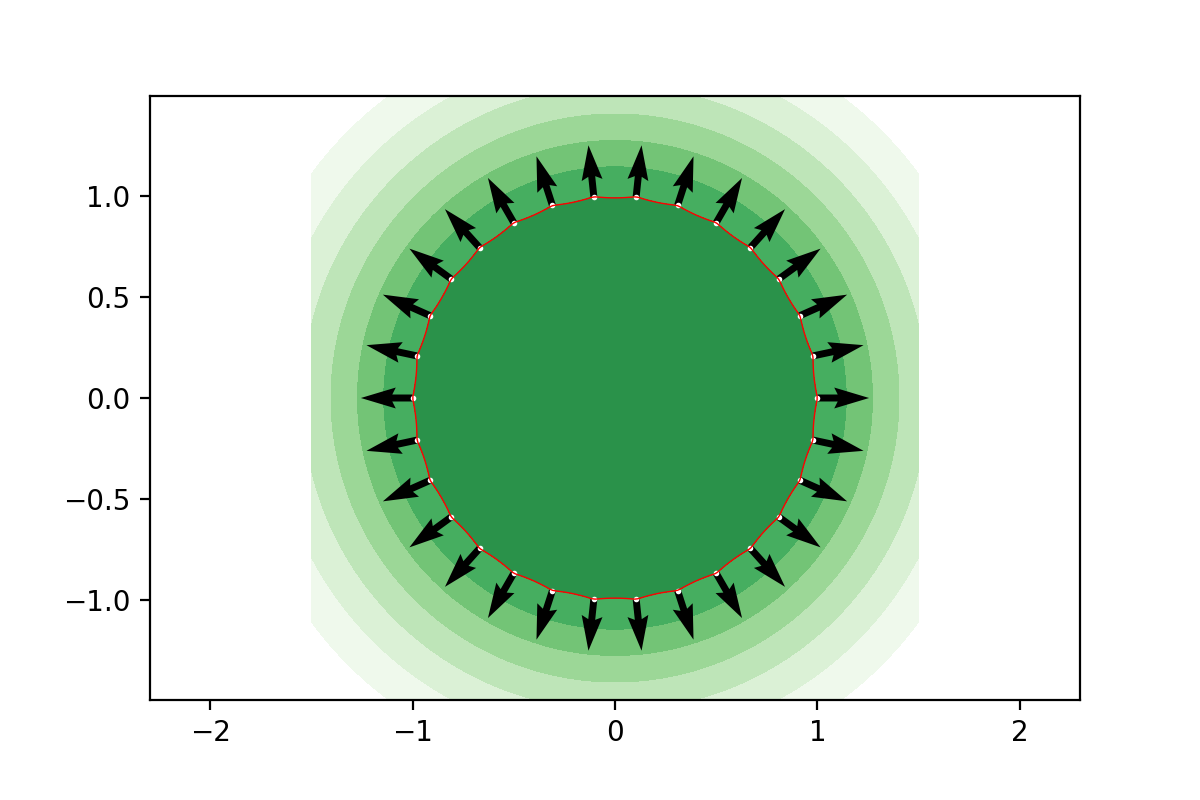}
  \caption{The top row depicts level lines of the signature function
    of the 30 white points and the implied normals obtained using the
    Gaussian kernel. The second row depicts the same for the Laplace
    kernel. The experiments in the first column correspond to no
    regularization ($\alpha=0$), whereas $\alpha=10^{-10}$ for the ones
    in the right column.}
  \label{fig:CircleNoNoise}
\end{figure}

First we consider an analytic curve, the unit
circle. We present the results of experiments using 30 regularly
spaced points along
the circle peformed with both the Gaussian and the Laplace kernel
without noise as well as with various degrees of noise for several
values of the regularization parameter $\alpha$. In the noisy
experiments, the percent is of the maximal distance between
consecutive points. In this way, the noise level depends only on
the data itself but can be related to the discretization level if
the data set represents a sample of points on the circle. The noise
itself is uniform in direction and in size. Figure
\ref{fig:CircleNoNoise} depicts the 
implied level line (in red) and the implied normals. The latter are
evaluated at the given points but, clearly, could be evaluated
anywhere. The level lines are computed by evaluating the data
signature function on a regular grid of a surrounding box. There is no
discernible difference between the two regularization levels,
$\alpha=0,10^{-10}$ in this smooth example with exact data. We shall
see later that the curvature is accurately captured with the Gauss
kernel, while it is not, for this coarse grid, for the Laplace
kernel. The use of regularization, however, limits the achievable
accuracy as reflected in an upcoming numerical estimation of
curvature below.  It can be seen in the images of Figure
\ref{fig:CircleNoNoise} that the signature function's level set (in
red) has sharp corners at the data points when using the
Laplace kernel. Away from the data, the level sets are smooth and
reflect the symmetry properties of the point cloud.

\subsection{A Square} Next we consider a non-smooth curve, the boundary of a square. This
example shows how the Laplace kernel can more closely follow the boundary as
it generates a solution of a PDE that is merely (H\"older)
continuous. It is interesting to observe how the level lines of the signature
function obtained with the Gauss kernel can also reproduce the
square but since the signature function is analytic it does so by
splitting the whole curve into two smooth closed curves\footnote{This
  is the reason why the normal points into the square in this
  case}, of which union the square is a part of. We shall soon see that
this does not prevent the use of the 
Gauss kernel in more general circumstances but leads to the need of
using some regularization to control the shape of the level lines. Again it is
apparent how the level lines smooth out and simplify away from the
non-smooth curve considered.
\begin{figure}
  \includegraphics[scale=.45]{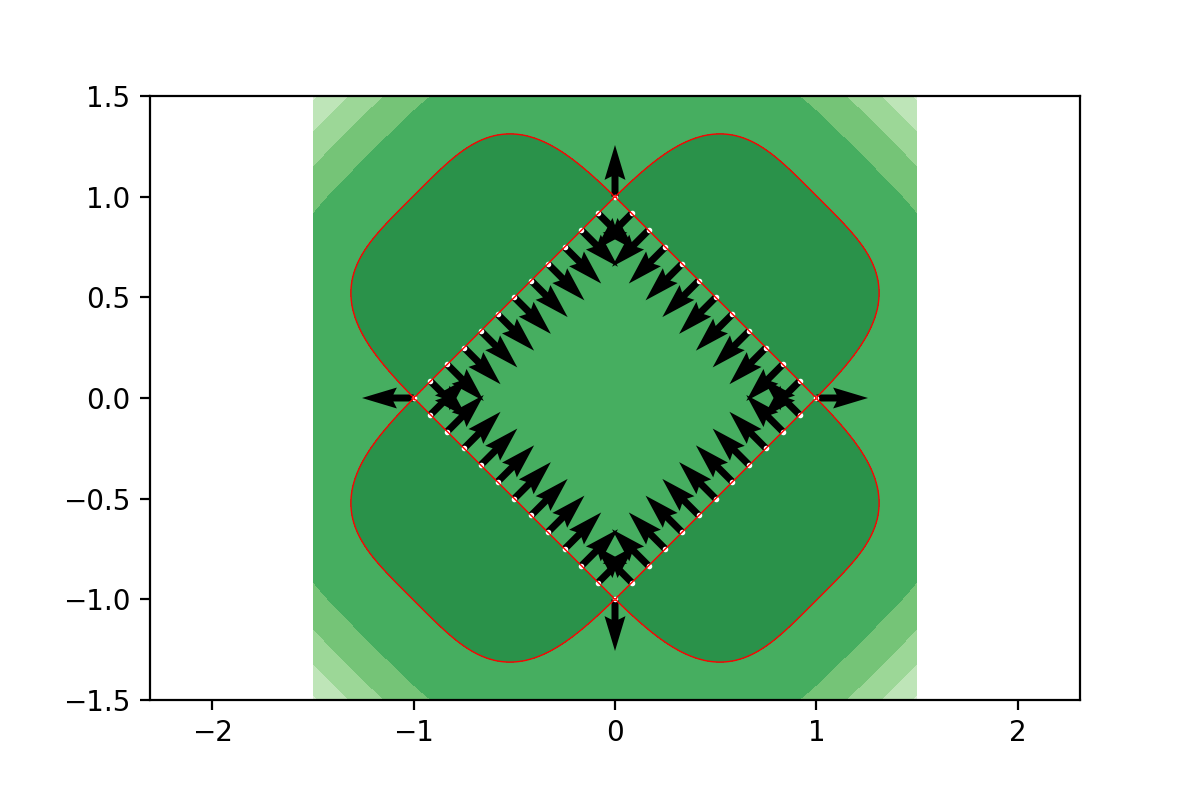}
  \includegraphics[scale=.45]{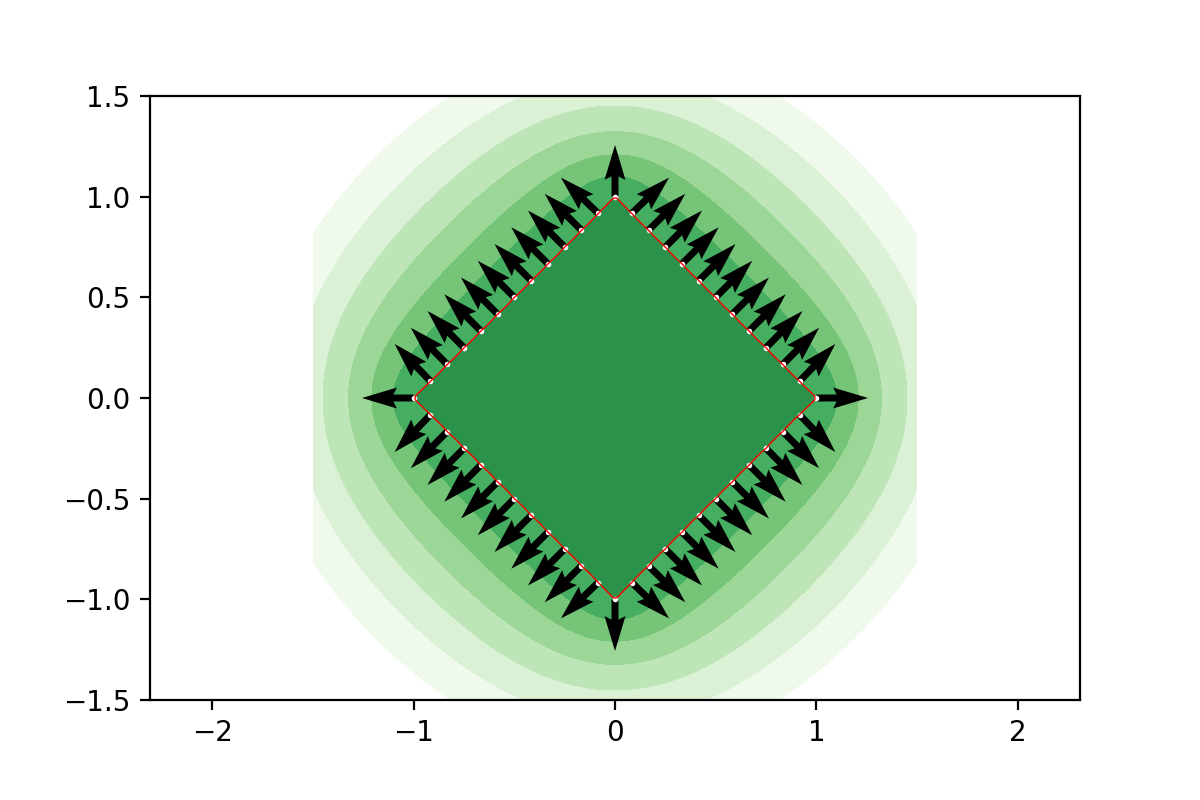}
  \caption{The level lines of the signature function
    of the 48 white points and the implied normals obtained using the
    Gaussian kernel (left) and Laplace Kernel (right). We used
    $\alpha=10^{-10}$ for both.}
  \label{fig:SquareNoNoise}
\end{figure}
\subsection{A sector} In Table \ref{table:sectorAccuracy} we
compute the error between the exact and implied curvatures for a
sector of a unit circle of angle $\frac{\pi}{16}$ for both the Gauss
and Laplace kernels. It shows the high accuracy achieved with
relatively few points by the Gauss kernel. It also showcases the
ability of the (regularized) Laplace kernel to obtain a good
approximation. We restrict the error analysis to the middle quarter of
the sector since the accuracy naturally deteriorates in the regions around the
end points. It is also apparent that the accuracy of the unregularized
Gauss kernel saturates (due to the increasing ill-conditioning of the
discrete system), while regularization is essentially unnecessary for
the Laplace kernel, again due to the fact that the discrete system
corresponds to a differential operator of ``minimal order'' and leads
to significantly less ill-conditioning.
\begin{center}
\begin{table}[!htbp]
\begin{tabular}{|l|l|l|c|c|c|c|}\hline
  \multicolumn{2}{|l|}{Method/Points} & 32 &64 & 128 &256 \\\hline
  Gauss&$\alpha=0$ & $2.26\times 10^{-6}$&$2.16\times 10^{-6}$&$1.54\times 10^{-6}$& $5.51\times 10^{-6}$\\\hline
  &$\alpha=10^{-10}$ &$5.47\times 10^{-4}$ &$2.78\times 10^{-4}$&$1.24\times 10^{-4}$&$4.22\times 10^{-5}$\\\hline
  Laplace&$\alpha=0$ & $4.84$&$0.24$ &$5.12\times 10^{-4}$& $1.01\times 10^{-6}$\\\hline
  &$\alpha=10^{-10}$ & $4.84$&$0.24$ &$5.12\times 10^{-4}$&$7.55\times 10^{-7}$\\\hline
\end{tabular}
  \vspace{2pt}
  \caption{The average relative error between the exact and the implied
  curvature at the points found in the middle quarter of a sector of
  the unit circle of aperture $\frac{\pi}{16}$.}
  \label{table:sectorAccuracy}
\end{table}
\end{center}
The experiments considered so far show the respective benefits of the two kernels. The
Gauss kernel delivers an analytic signature which attempts to place
the data points on a union of smooth curves, while the Laplace kernel
leads to a merely continuous signature which can therefore exhibit
level sets that more closely follow the data points, even as they go
through corners (singularities in general).
\subsection{Noisy Circle and Square} 
\begin{figure}
  \includegraphics[scale=.35]{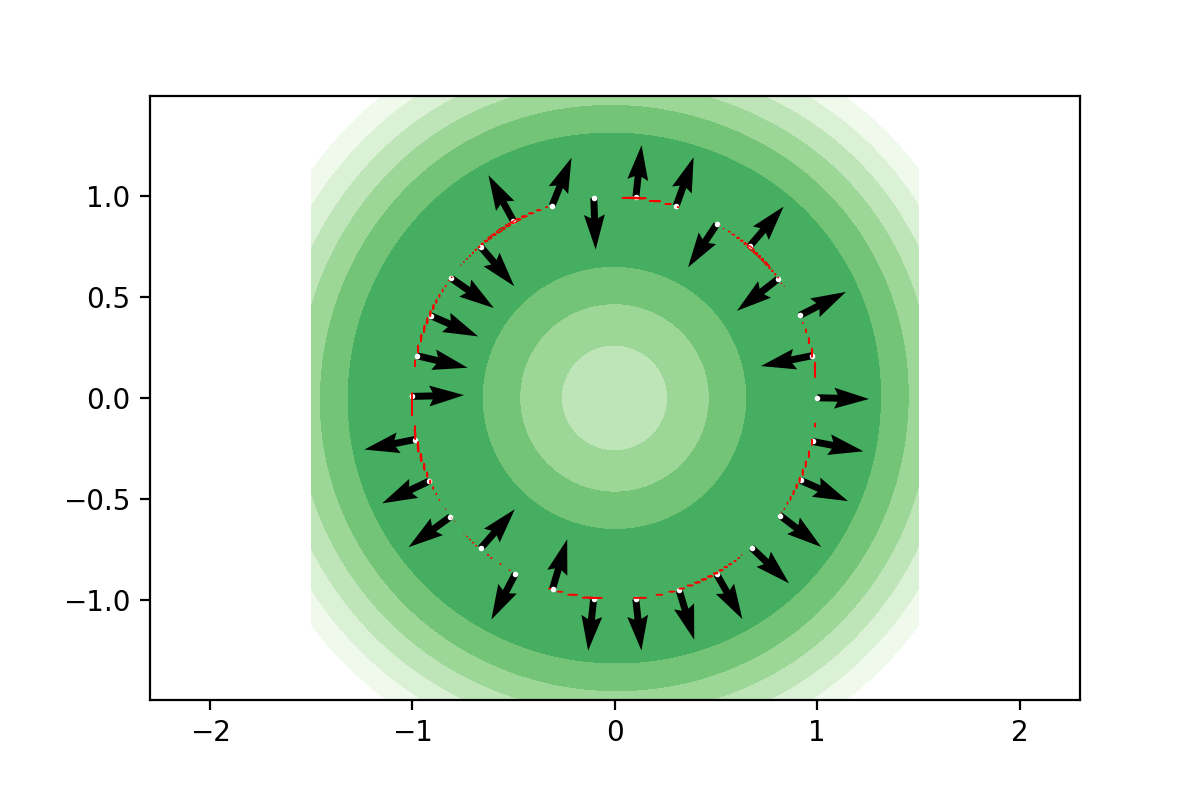}
  \includegraphics[scale=.35]{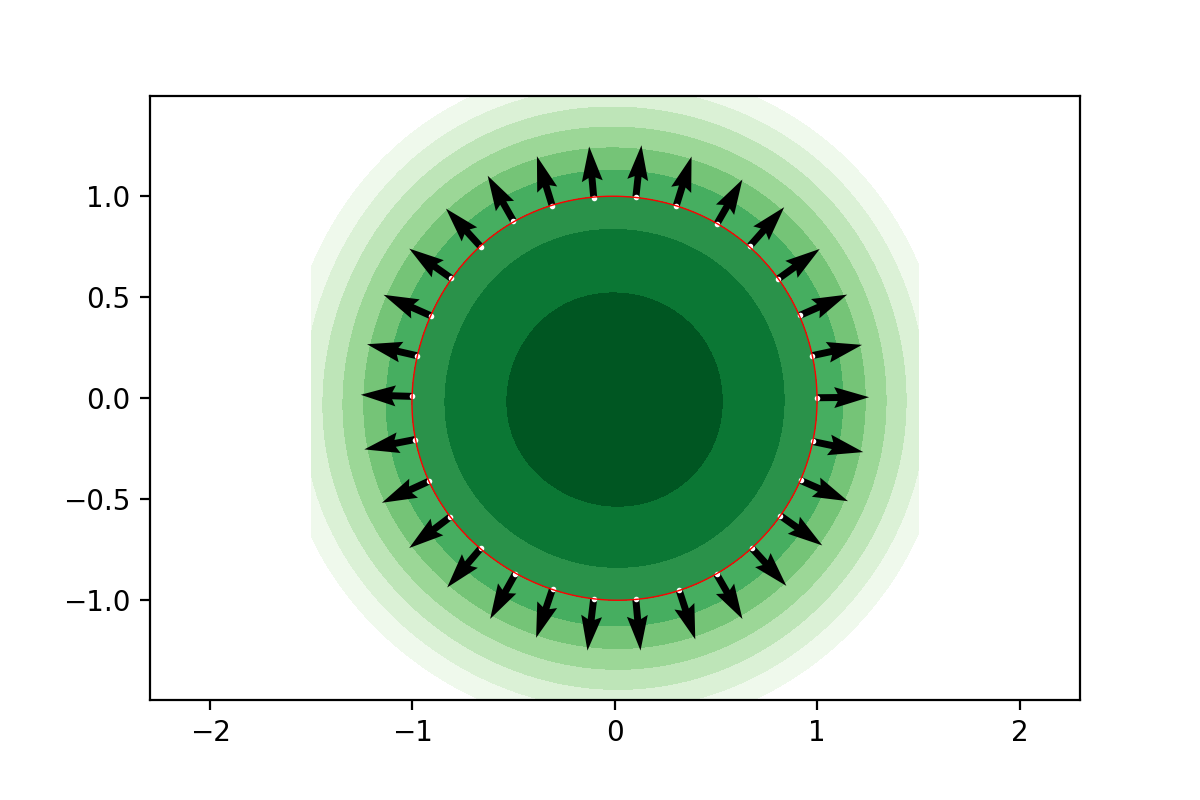}
  \includegraphics[scale=.35]{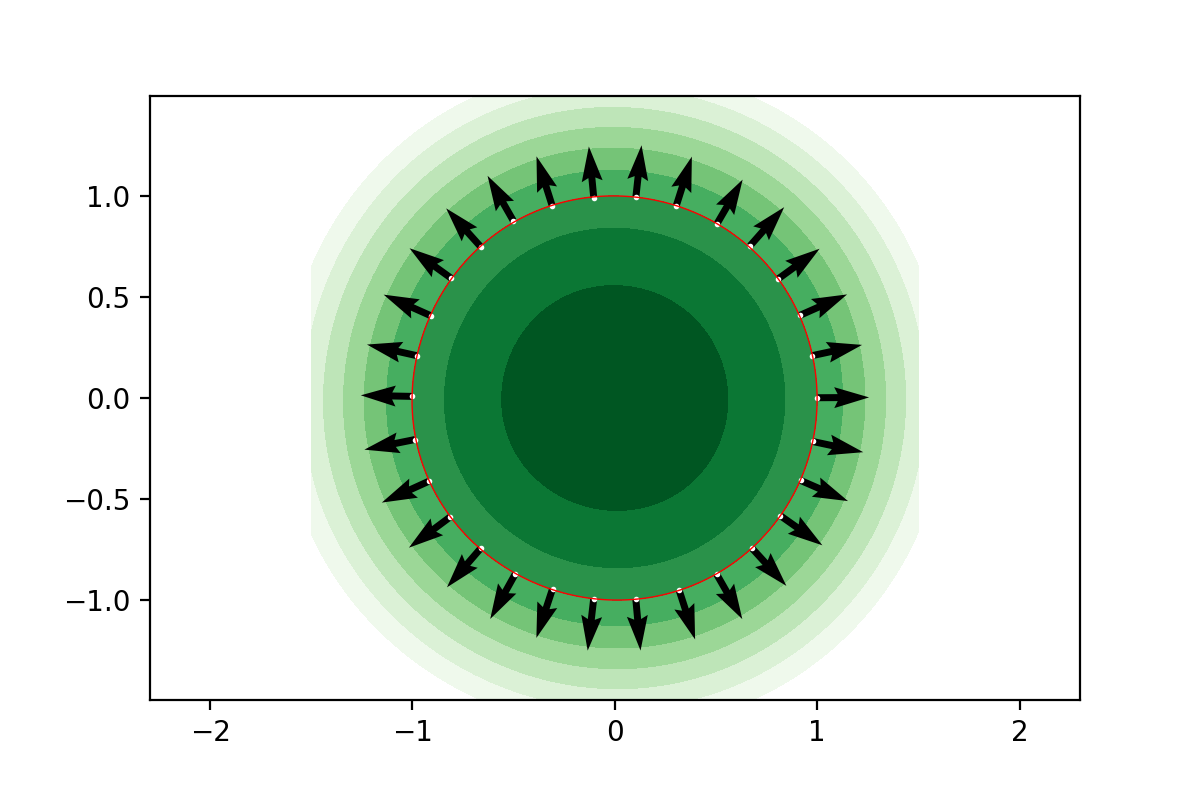}
  \caption{The circle example revisited by randomly displacing the
    original exact sample by 5\% noise. Depicted are the level lines
    and implied normals obtained by means of the Gauss kernel based
    signature function computed with different regularization levels
    $\alpha=0,.01,.1$ in increasing order from left to right.}
  \label{fig:circleNoise1}
\end{figure}

\begin{figure}
  \includegraphics[scale=.35]{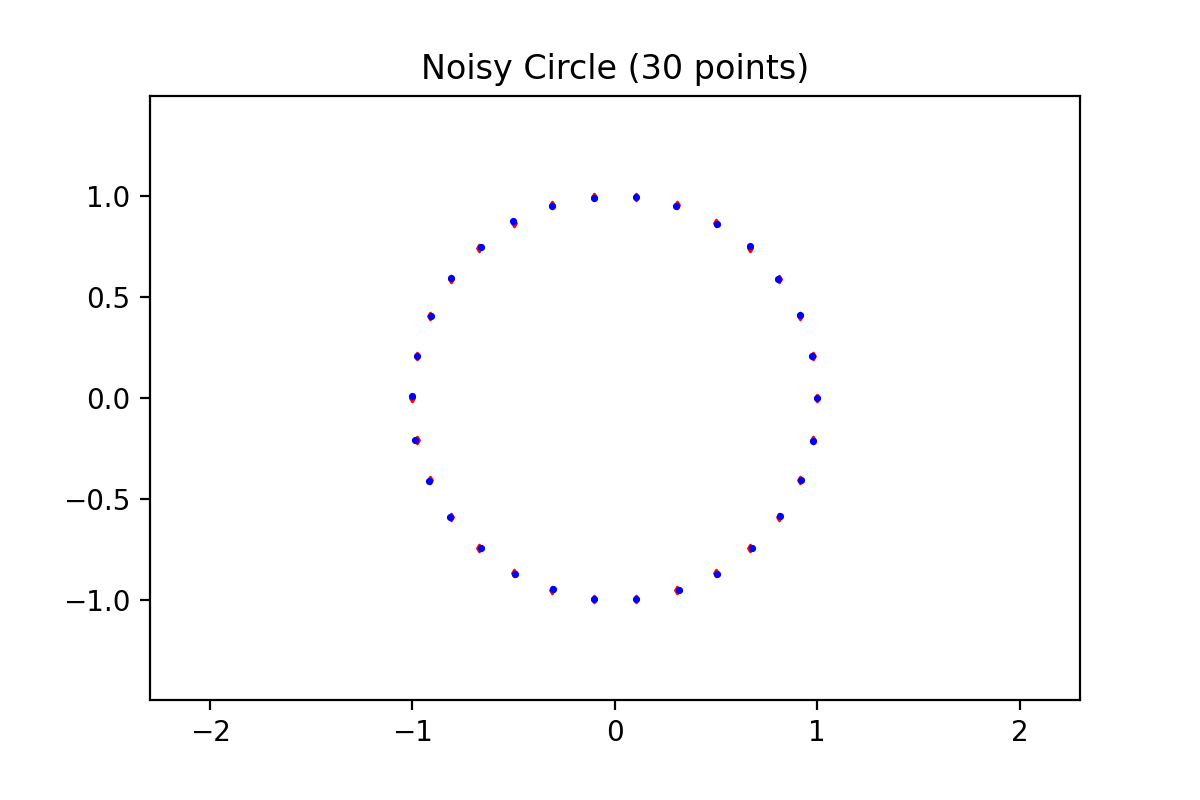}
  \includegraphics[scale=.35]{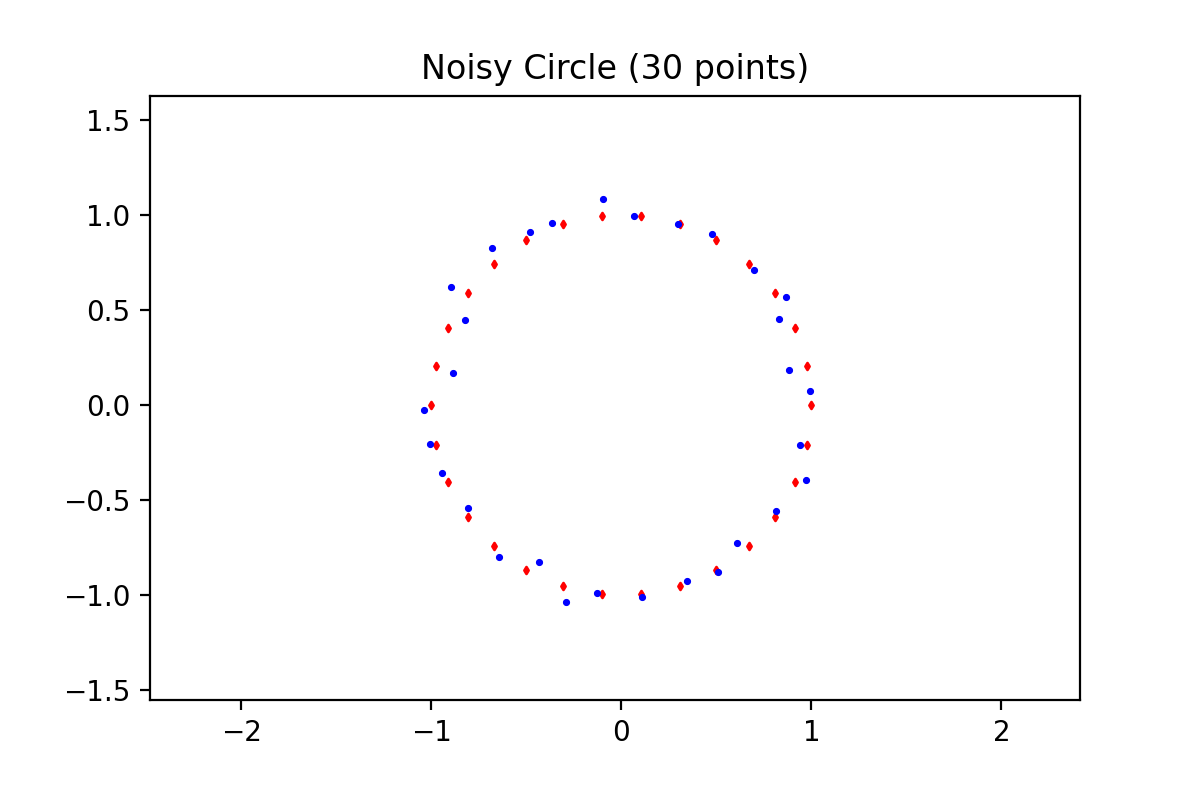}\\
  \includegraphics[scale=.35]{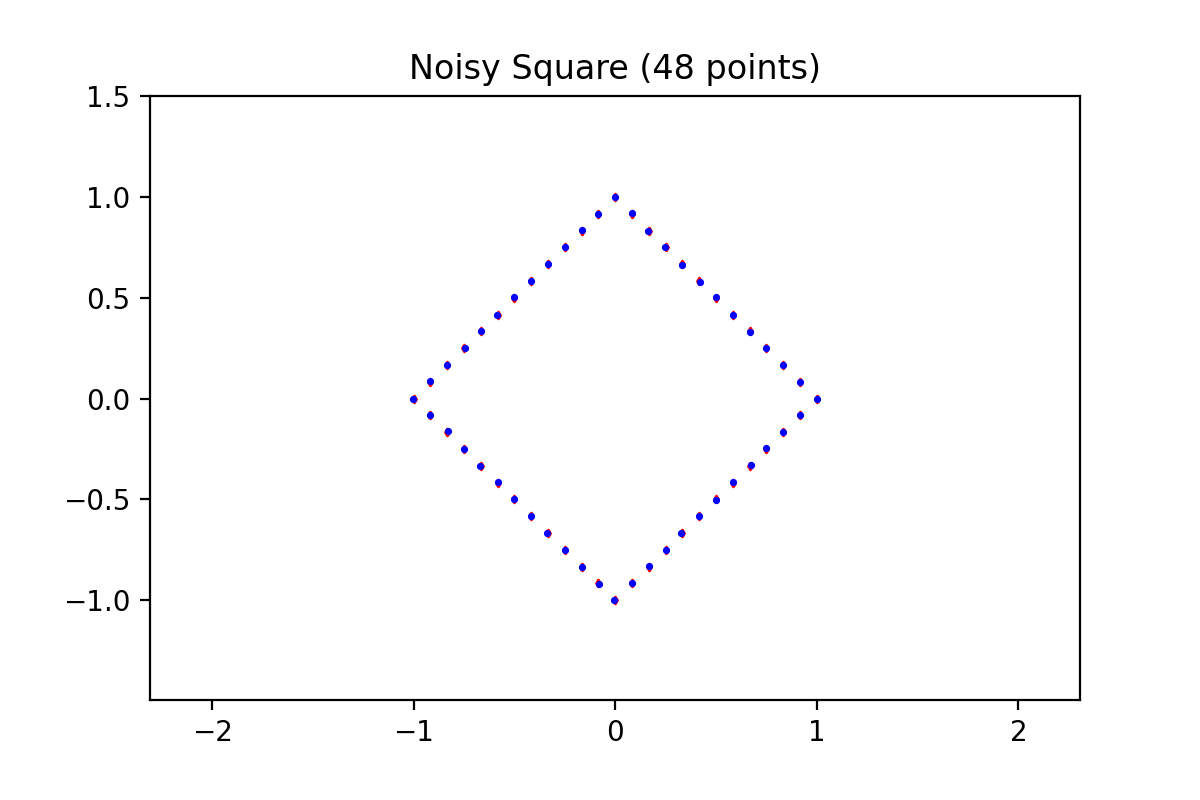}
  \includegraphics[scale=.35]{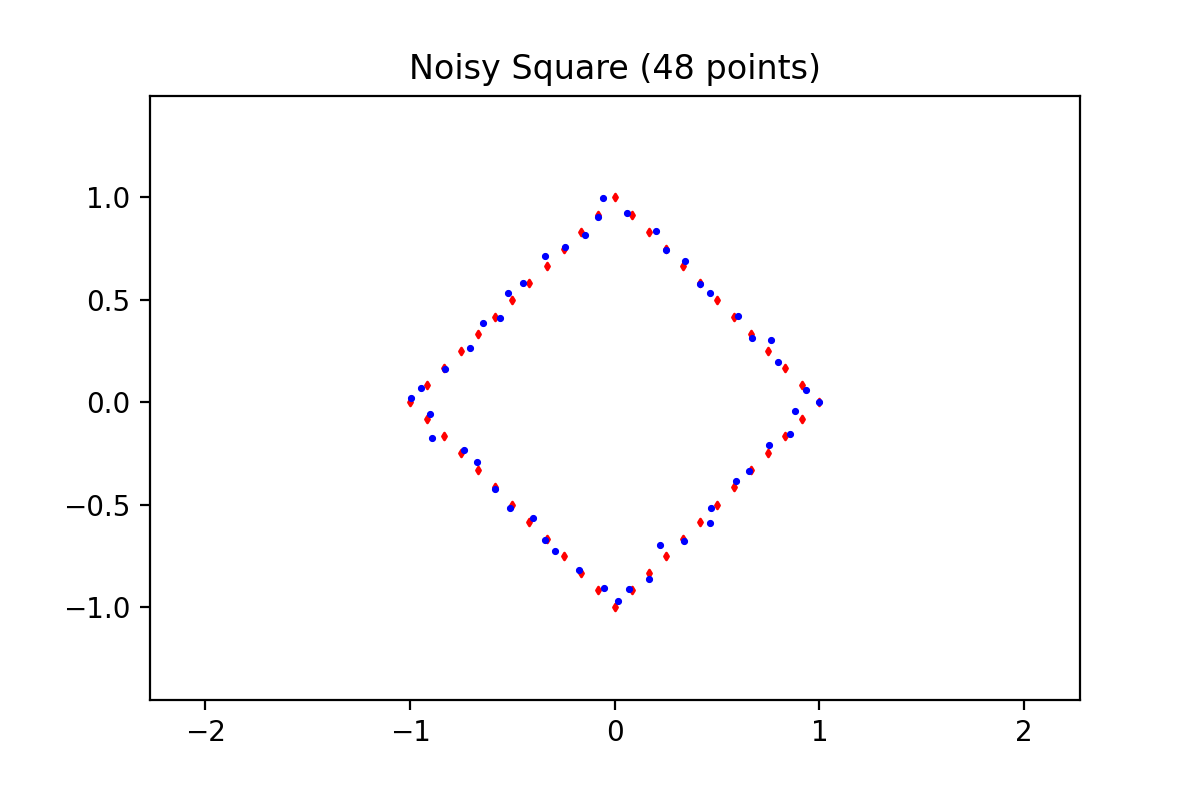}
  \caption{The original set of data points is shown in all images as
    red diamonds. The blue dots show the noise-perturbed data points
    at the different noise levels: 5\% and 50\% from left to right. }
  \label{fig:noisyShapes}
\end{figure}

\begin{figure}
  \includegraphics[scale=.35]{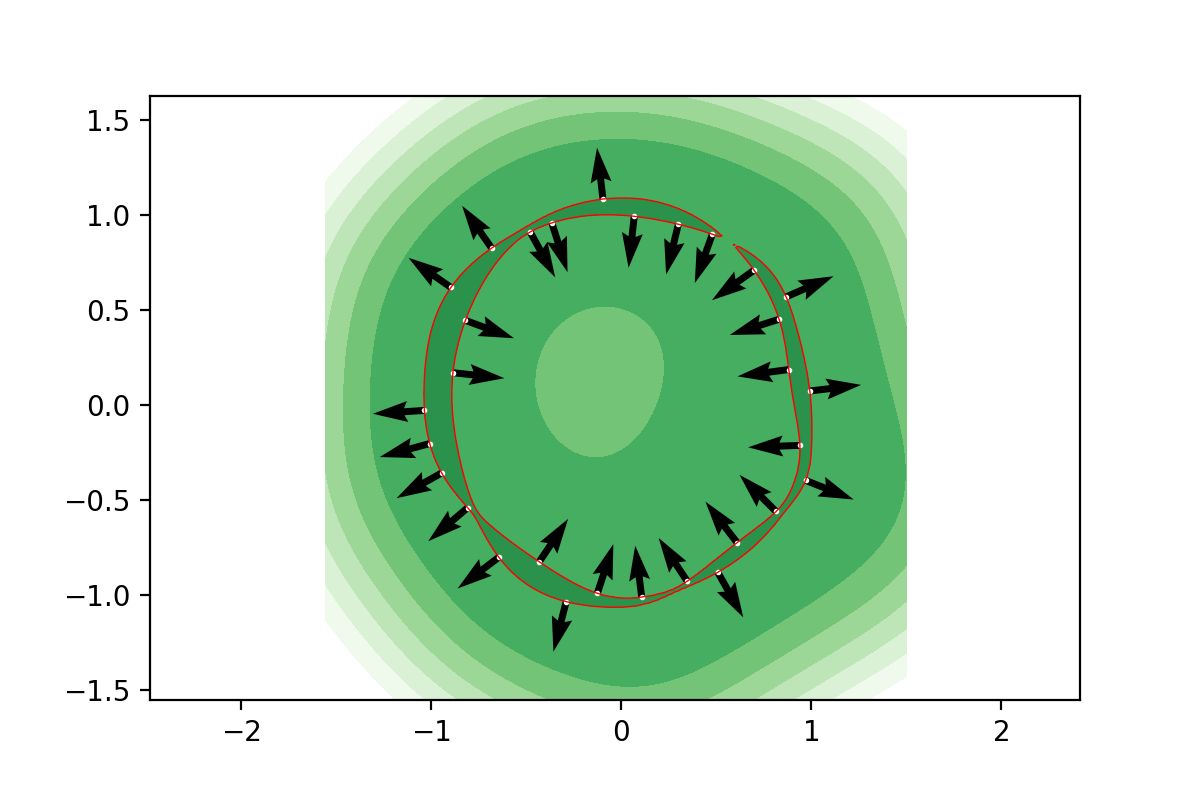}
  \includegraphics[scale=.35]{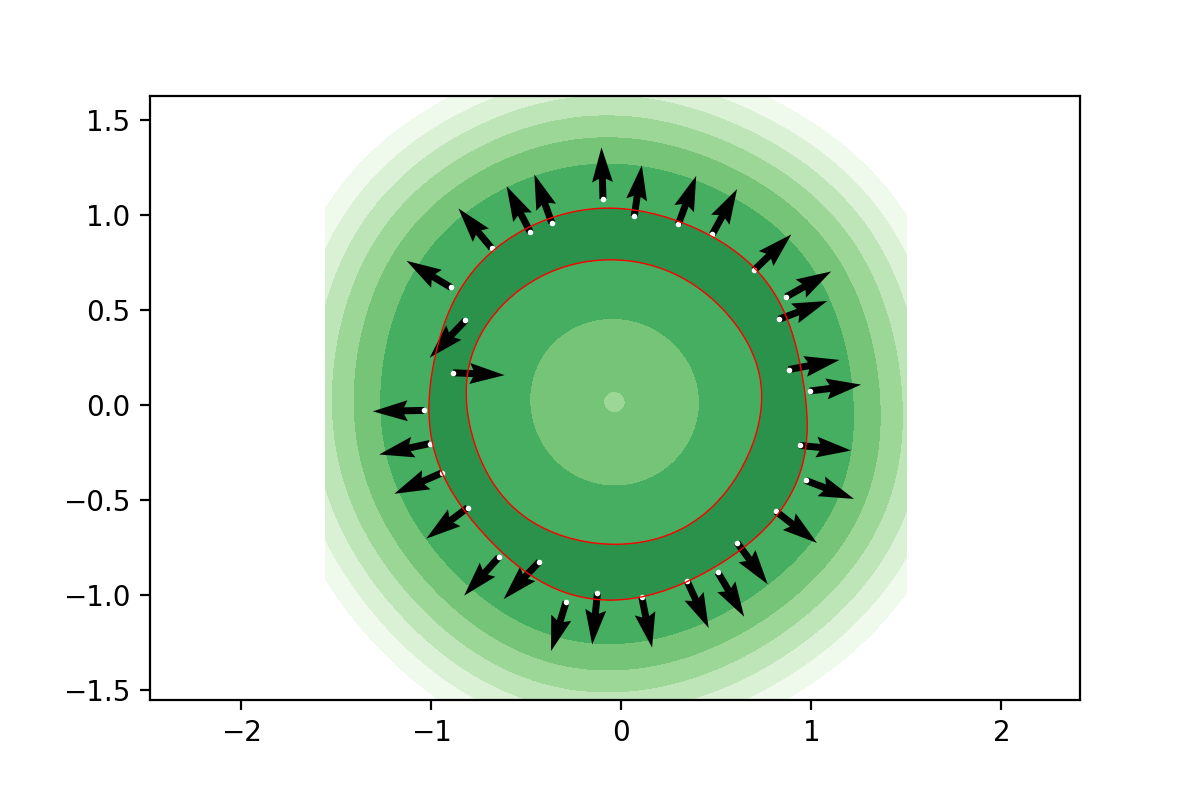}
  \includegraphics[scale=.35]{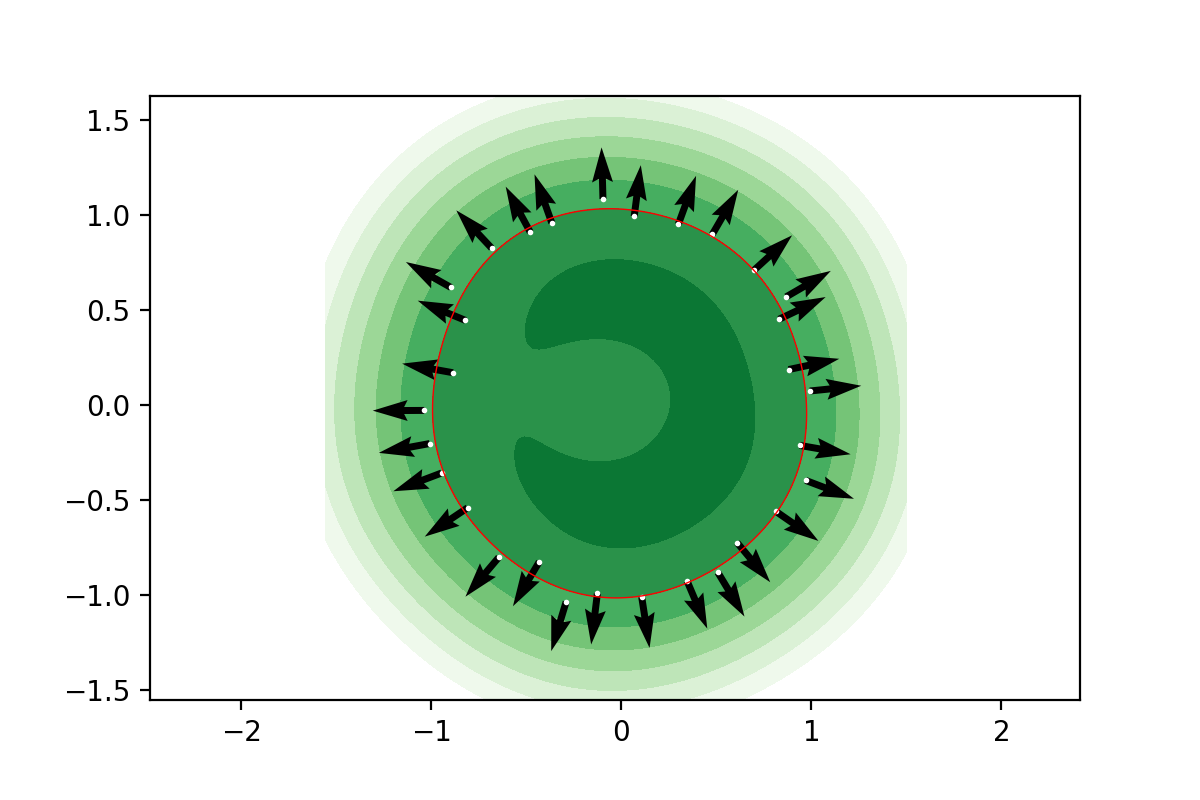}
  \caption{The circle example revisited by randomly displacing the
    original exact sample by 50\% noise. Depicted are the level lines
    and implied normals obtained by means of the Gauss kernel based
    signature function compute with different regularization levels
    $\alpha=0,.01,.1$ in increasing order from left to right.}
  \label{fig:circleNoise3}
\end{figure}

\begin{figure}
  \includegraphics[scale=.35]{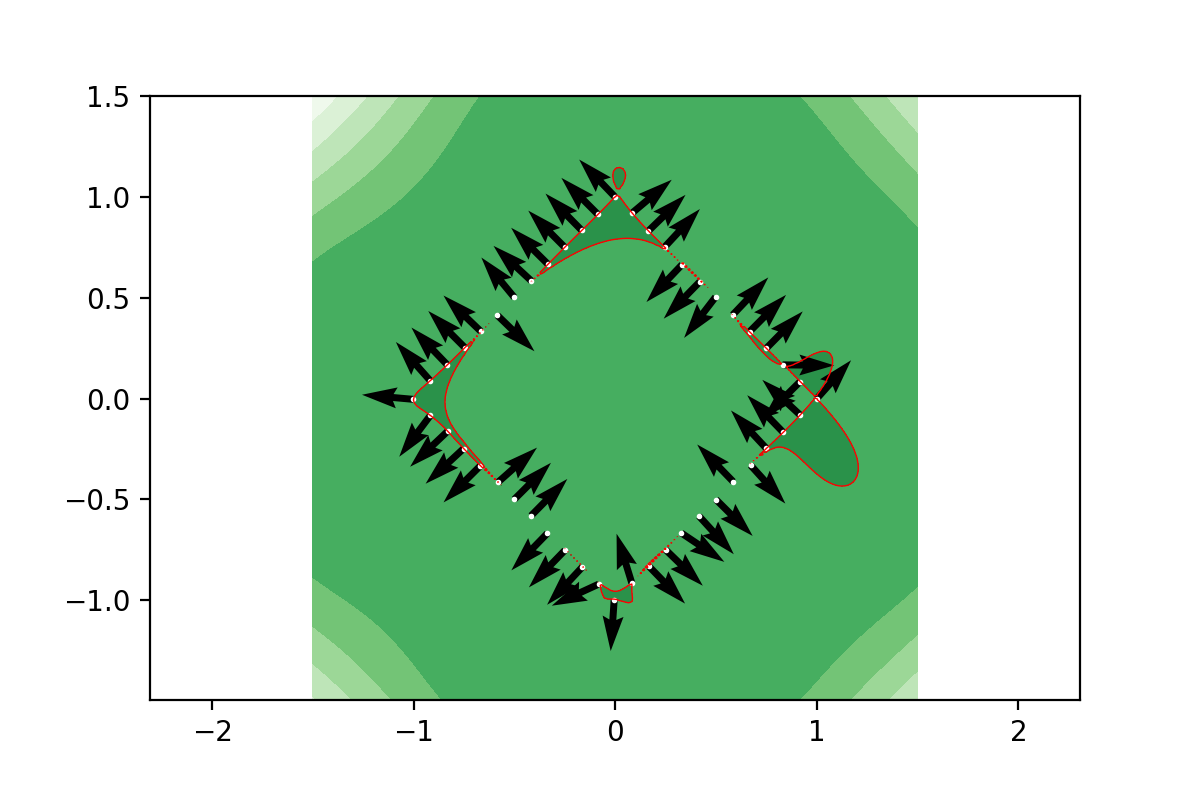}
  \includegraphics[scale=.35]{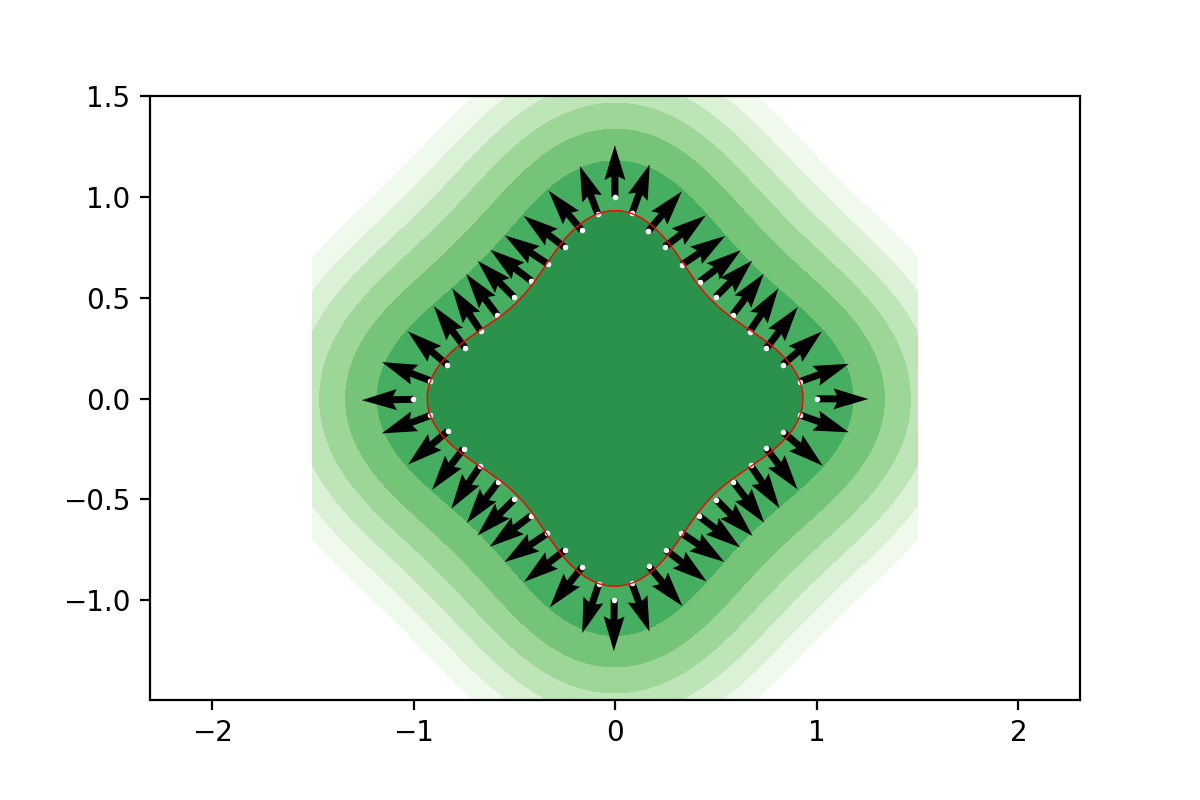}
  \includegraphics[scale=.35]{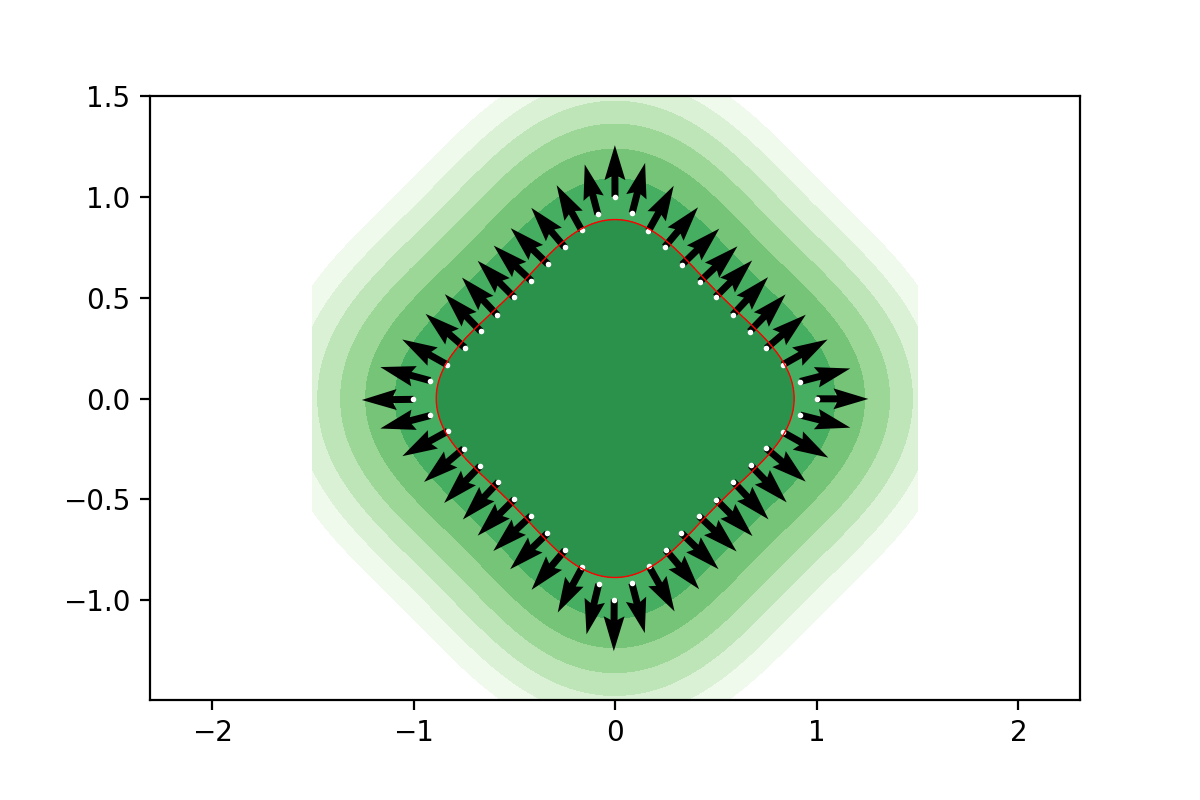}
  \caption{The square example revisited by randomly displacing the
    original exact sample by 5\% noise. Depicted are the level lines
    and implied normals obtained by means of the Gauss kernel based
    signature function computed with different regularization levels
    $\alpha=0,.01,.1$ in increasing order from left to right.}
  \label{fig:squareNoise1}
\end{figure}

\begin{figure}
  \includegraphics[scale=.35]{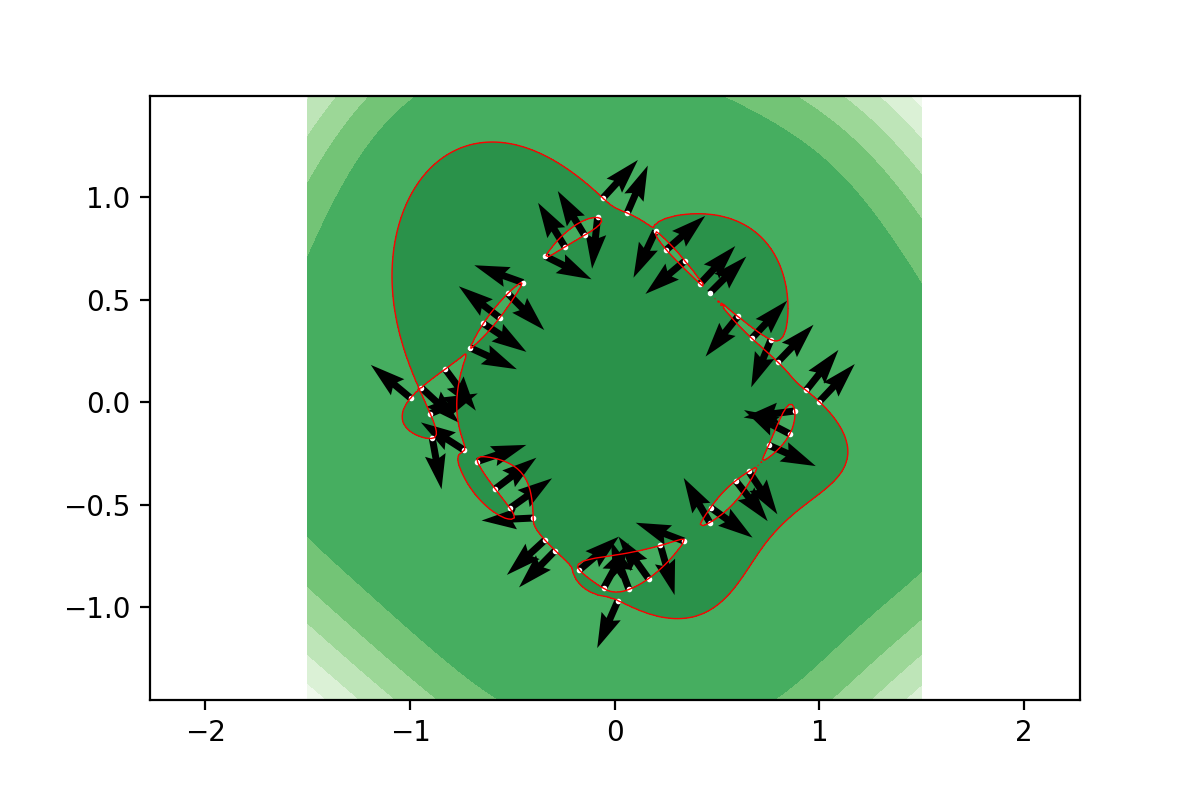}
  \includegraphics[scale=.35]{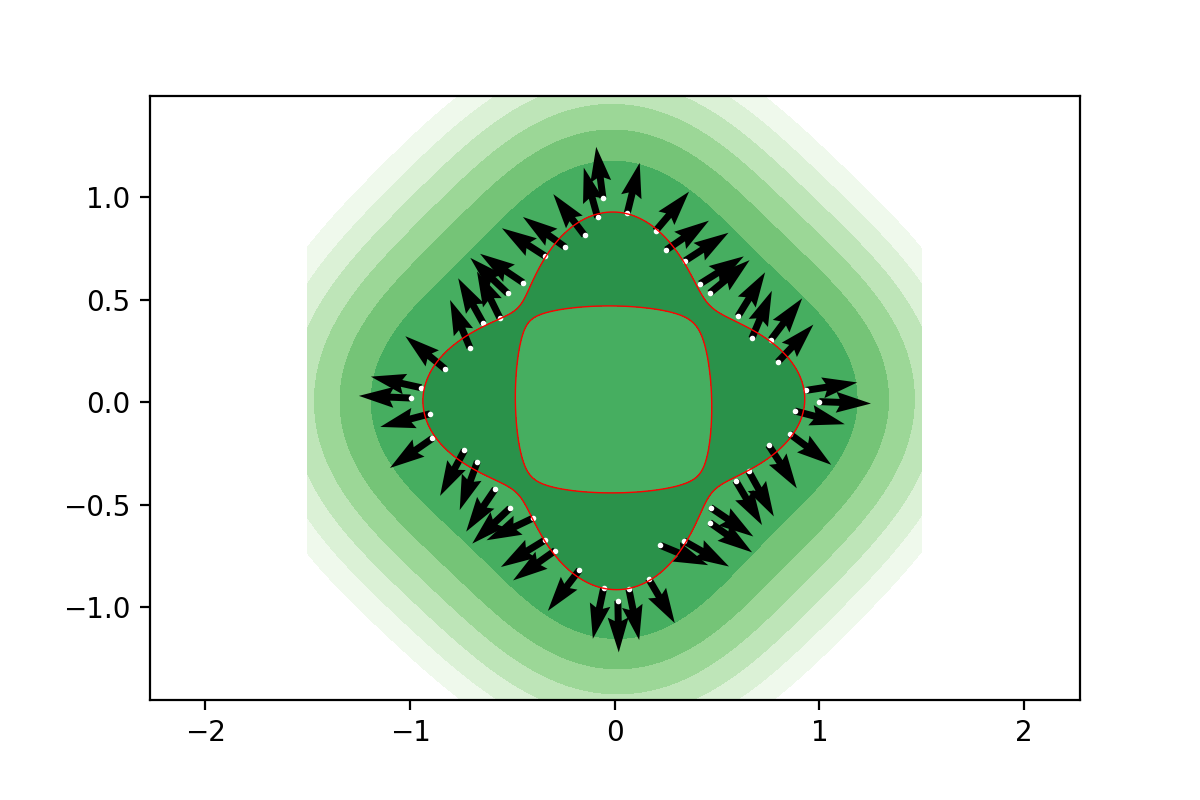}
  \includegraphics[scale=.35]{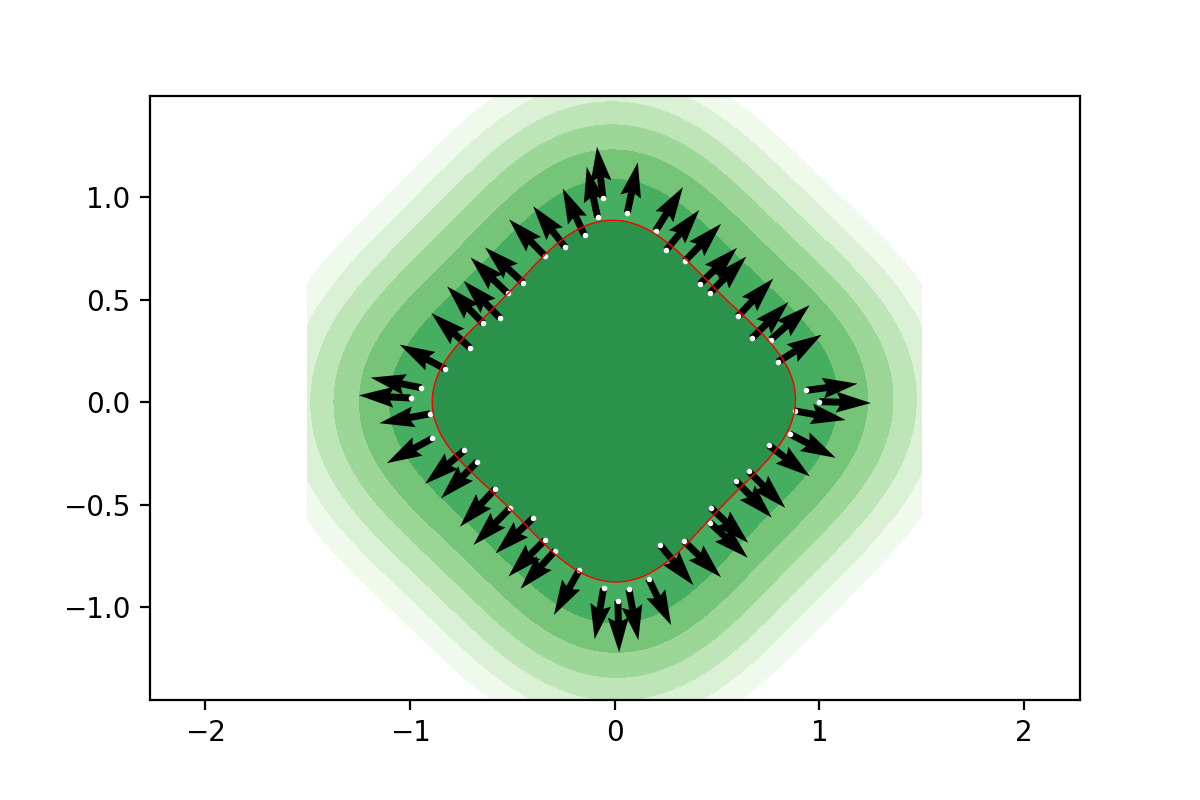}
  \caption{The square example revisited by randomly displacing the
    original exact sample by 50\% noise. Depicted are the level lines
    and implied normals obtained by means of the Gauss kernel based
    signature function computed with different regularization levels
    $\alpha=0.01,.1$ in increasing order from left
    to right}
  \label{fig:squareNoise3}
\end{figure}

The benefits just discussed are even more
evident in the experiments depicted in Figures
\ref{fig:circleNoise1}-\ref{fig:circleNoise3} and
\ref{fig:squareNoise1}-\ref{fig:squareNoise3}, where the data points
are randomly displaced from their original position along the circle
and along the boundary of the square. The new points are obtained from
the old ones by adding a displacement of 5\% and 50\% which is
uniform in direction and in distance. The percentage refers to maximal
displacement size as a fraction of the maximal distance between
consecutive data points. In this way, the error is directly related to
the data set, which is in general all that is available. Notice,
however, that the ordering (parametrization) of the points is not
in any way required knowledge for the construction of the signature
function. Figure \ref{fig:noisyShapes} shows the realization of the
random data sets with different levels of noise for the experiments
depicted in Figures \ref{fig:circleNoise1}-\ref{fig:circleNoise3} and
\ref{fig:squareNoise1}-\ref{fig:squareNoise3} obtained by the use of
the Gaussian kernel. The corresponding experiments with the Laplace
kernel are based on a different realization of the perturbed data set
that is not depicted in this paper.
It is evident how the analytic signature tries to accomodate smooth
level lines through the data without necessarily connecting the points
in what the human eye would consider the natural way. Notice, however, how
the implied normals can still approximately capture the tangent line to the
``average curve''. The introduction of regularization, by allowing for
approximate interpolation, makes it possible for the method to
successfully connect the dots into a smooth line and capture its
normal vector. In these and subsequent numerical experiments, the level
line depicted in red is that corresponding to the average value of the
signature function on the data set.


\subsection{A Graph} 
\begin{figure}
  \includegraphics[scale=.4]{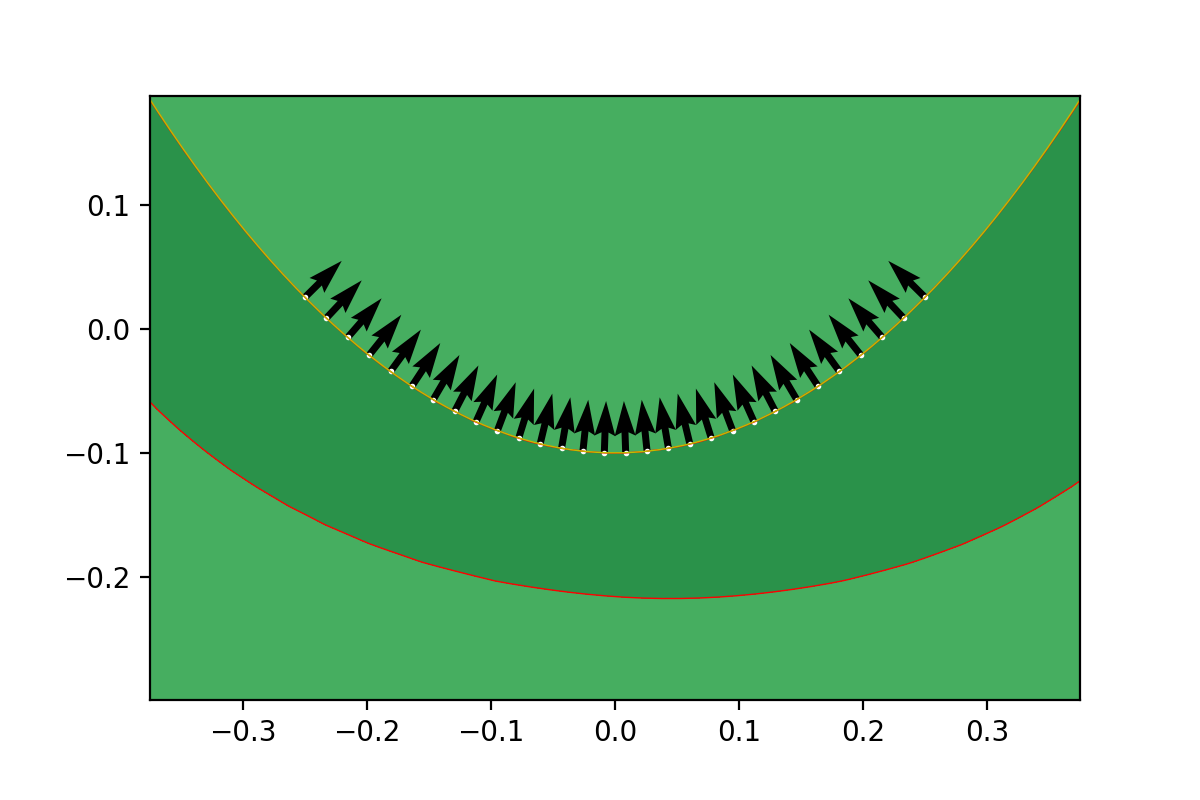}
  \includegraphics[scale=.4]{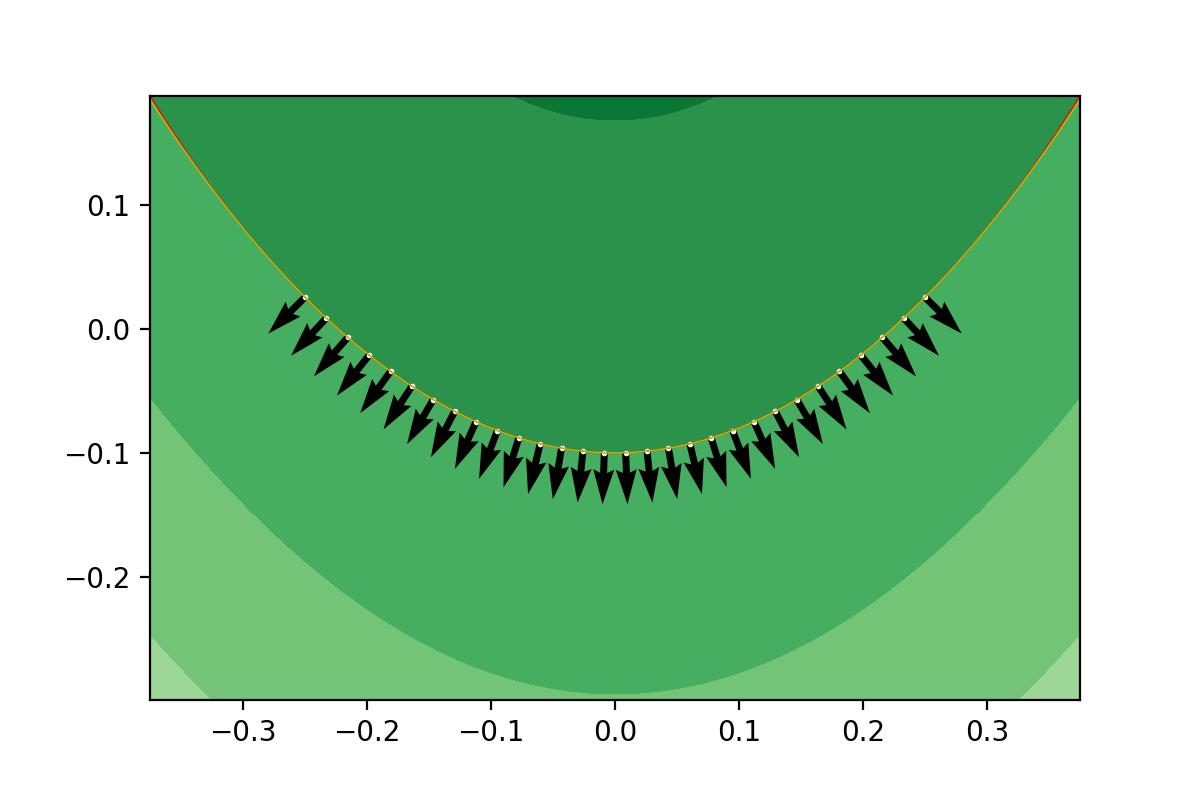}\\
  \includegraphics[scale=.4]{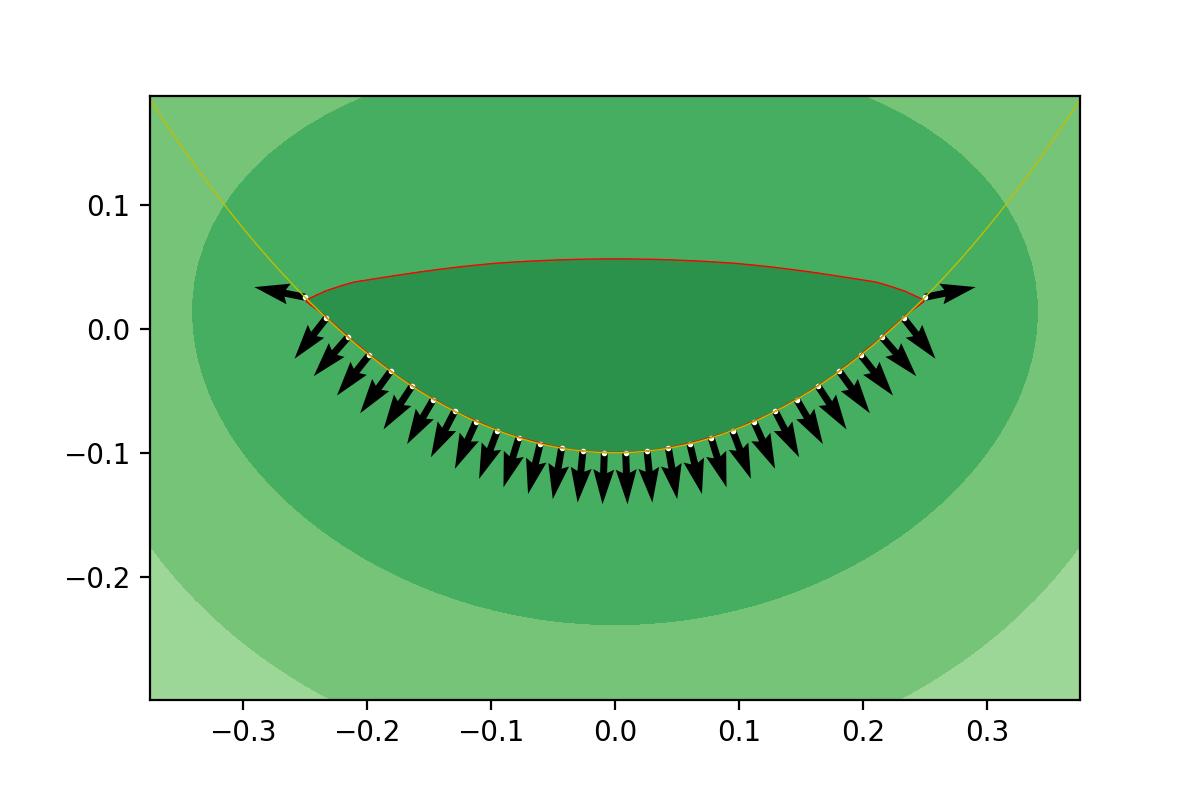}
  \includegraphics[scale=.4]{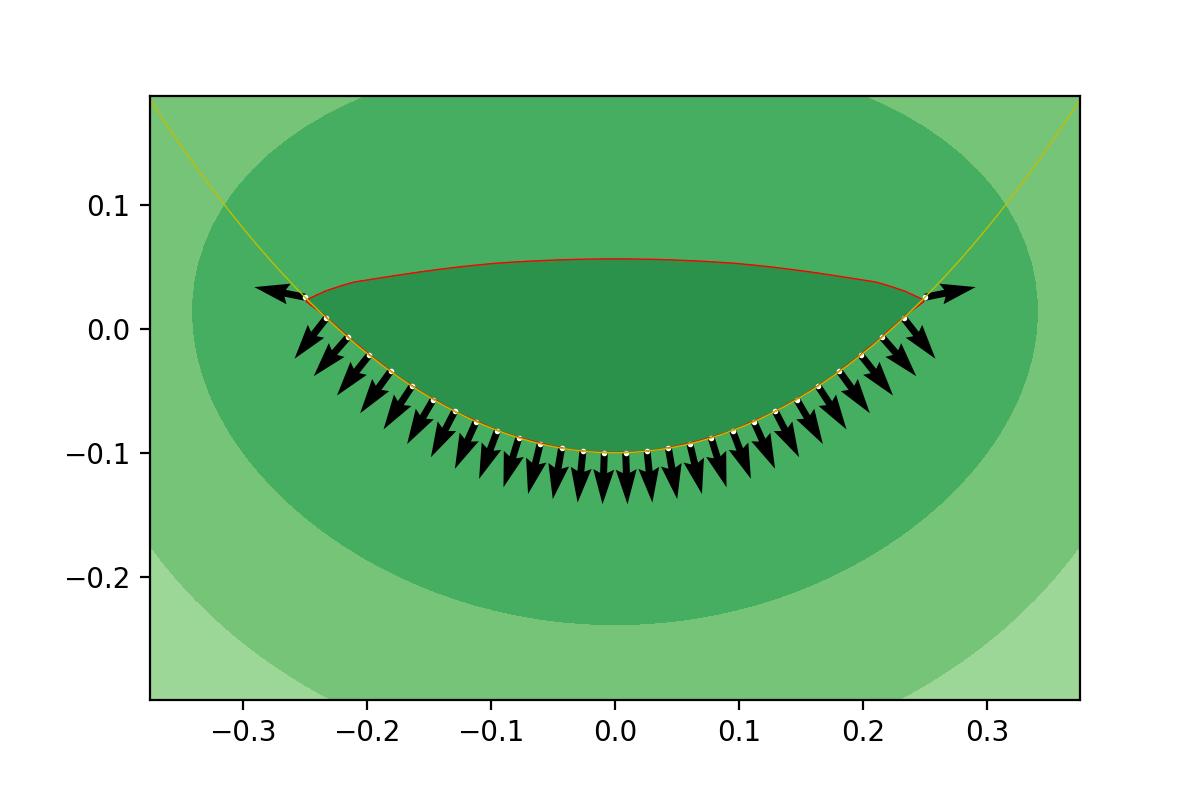}
  \caption{The top row depicts the implied level line (in red) of the signature function
    of the graph (in yellow) of the function $y=-1+x^2+|x|^{2.5}$ for
    $x\in[-.25,.25]$ along  with the implied 
    normals (at the data points) obtained using the Gaussian
    kernel. The second row depicts the same for the Laplace 
    kernel. The experiments in the first column correspond to no
    regularization ($\alpha=0$), where $\alpha=10^{-10}$ for the ones
    in the right column.}
  \label{fig:graphNoNoise}
\end{figure}

\begin{figure}
  \includegraphics[scale=.4]{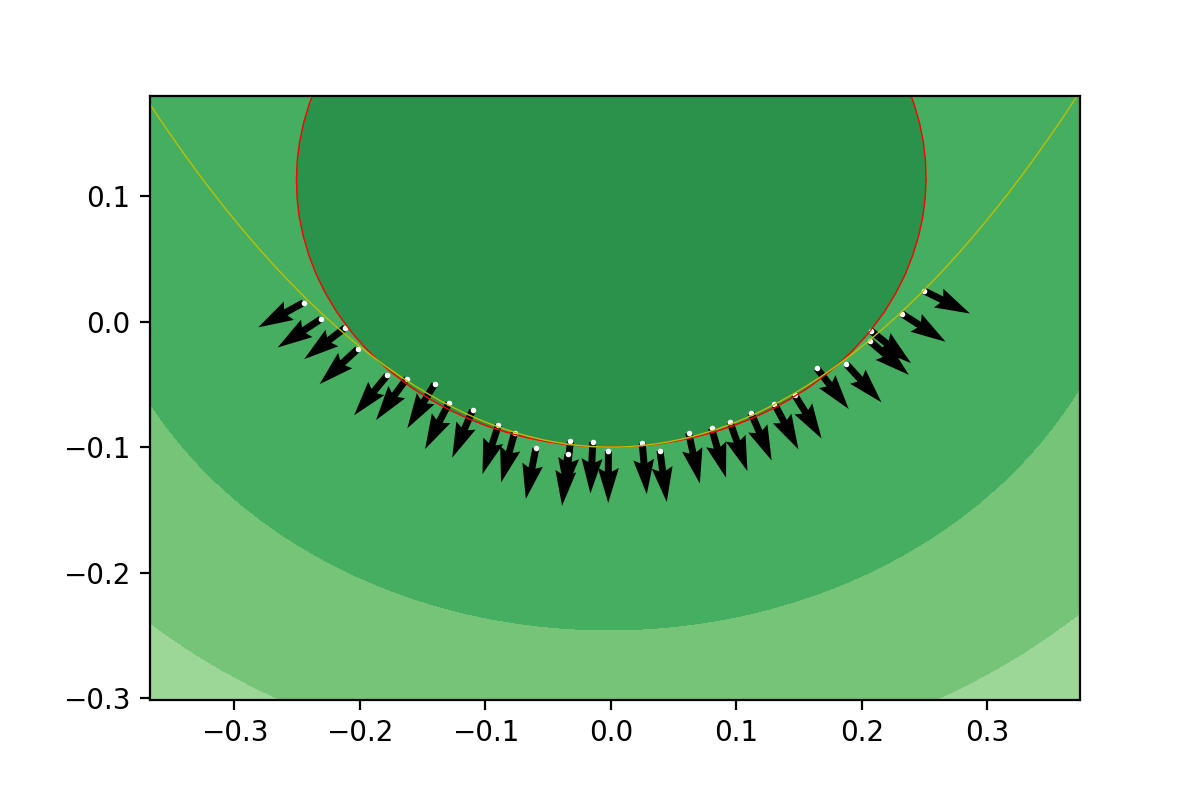}
  \includegraphics[scale=.4]{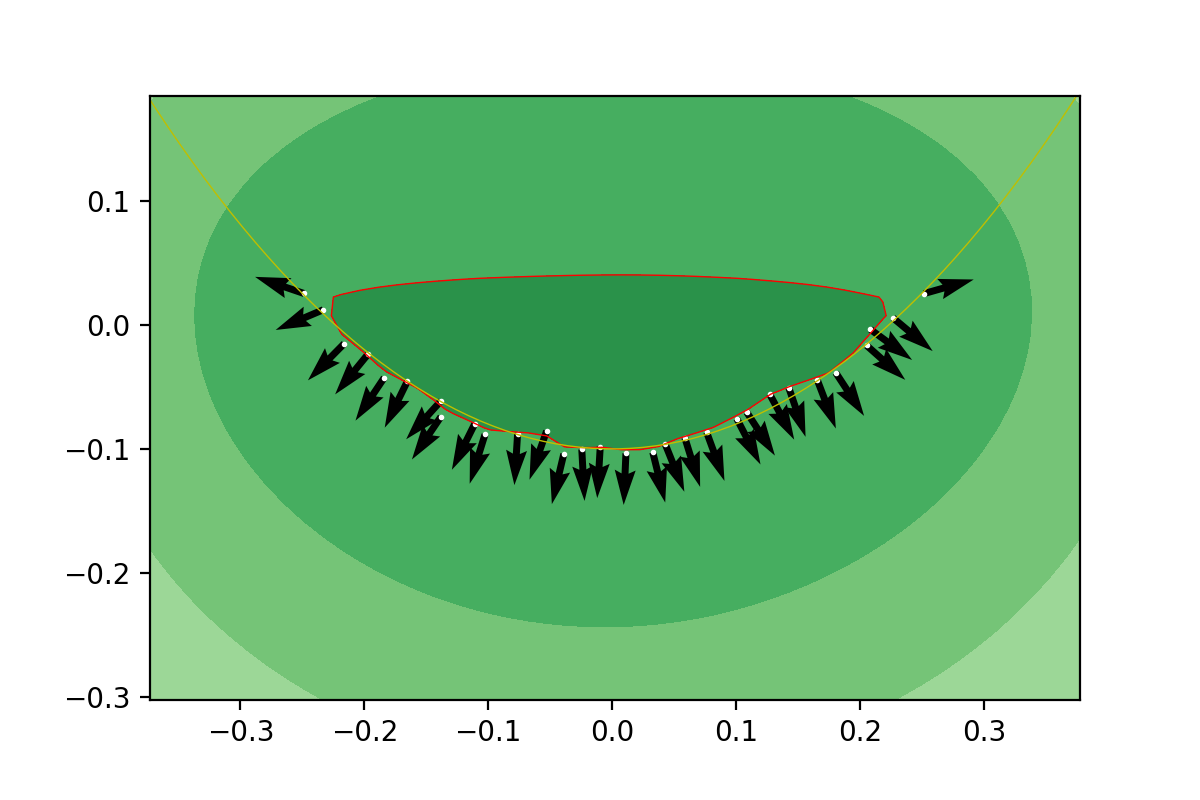}
  \includegraphics[scale=.4]{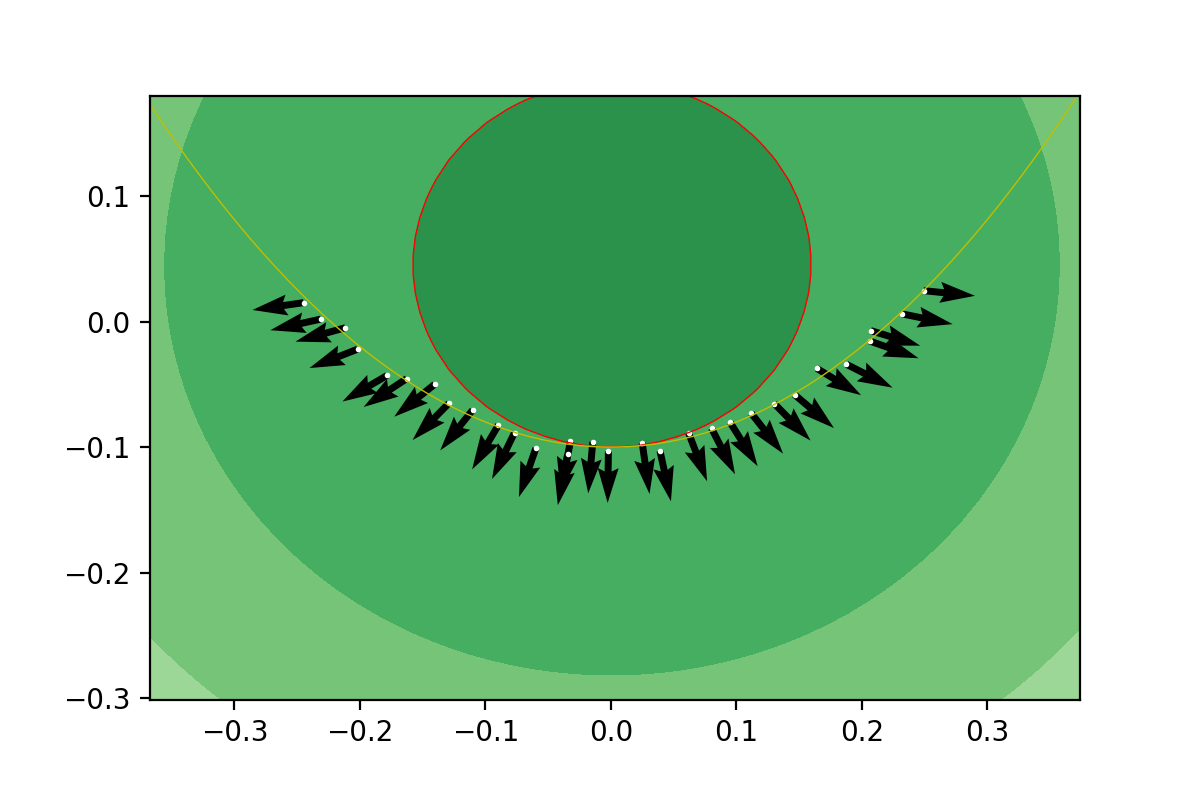}
  \includegraphics[scale=.4]{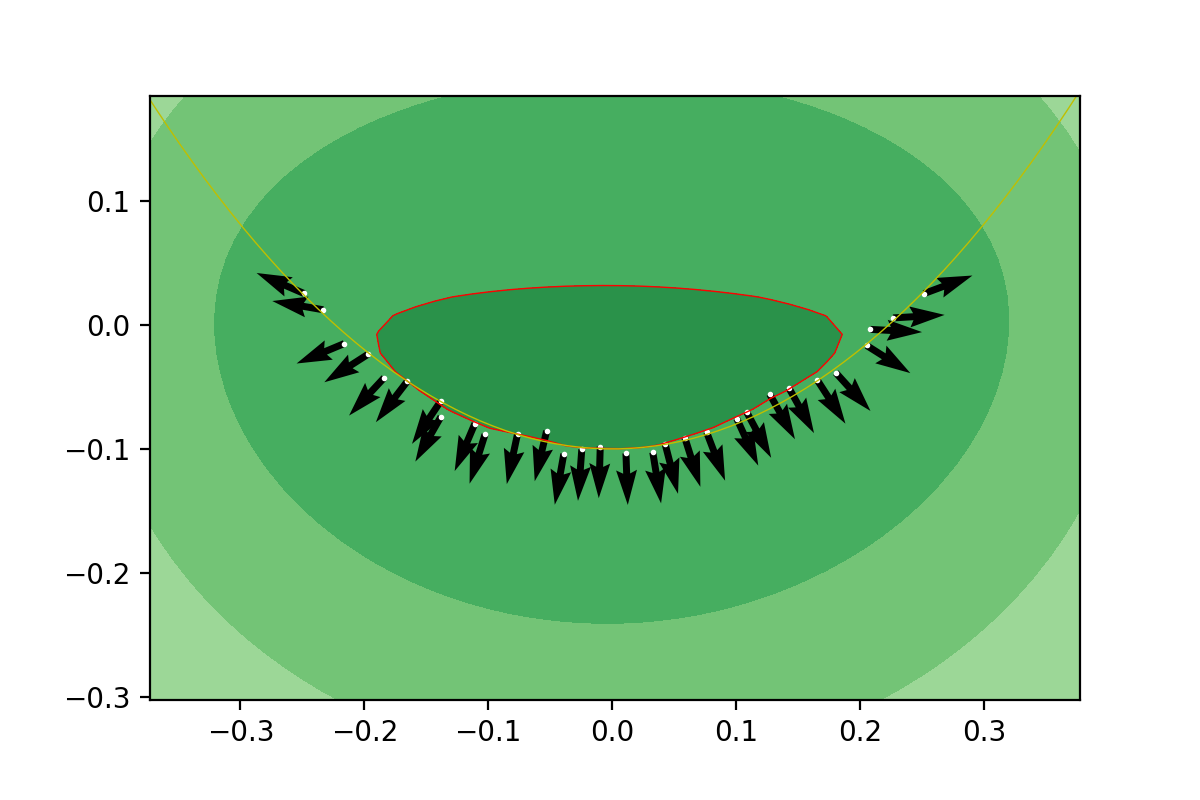}
  \caption{The graph example revisited by randomly displacing the
    original exact sample by 50\% noise. Depicted are the level lines
    and implied normals obtained by means of the Gauss (left column)
    and Laplace (right column) kernel based
    signature function computed with different regularization levels
    $\alpha=.01,.1$ in increasing order from top
    left to bottom right.}
  \label{fig:graphNoise}
\end{figure}

The experiments found in Figures \ref{fig:graphNoNoise}  and
\ref{fig:graphNoise} show the method applied to a non-smooth graph locally
around the origin without and with noise. In Tables
\ref{table:impliedNormal}-\ref{table:impliedCurvature} we record the
implied normals and curvatures computed at the origin with the Gauss
kernel at various regularization and noise levels. Again the normal
appears to be computed in a very stable manner across noise and
regularization levels.
\begin{center}
\begin{table}[h!]
{\tiny\begin{tabular}{|l|c|c|c|c|c|}\hline
  Noise/Regularization & 0 & $10^{-10}$ & .01 & .05 & .1 \\\hline
  0\% &$(-.000002,1.)$&$(0. ,-1.) $&
  &&\\\hline
  5\% &$(0.0568,-0.998)$&$(0.0025,-1.)$&$(-.000419,-1.)$&$(-.000368,-1.)$&$(-.000407,-1.)$\\\hline
  10\% &$(.000892,1.)$&$(-0.0152,-0.9999)$&$(.000821,-1.)$
   &$(.000332,-1.)$&$(.000835,-1.)$\\\hline
  50\% &$(0.016,1.)$&$(0.0885,0.996)$&$(0.0031 -1.)$&$(-.000419,-1.)$&$(-0.00364, -1.)$\\\hline
\end{tabular}}
  \vspace{2pt}
  \caption{Implied normal at the origin for the graph experiment.}
  \label{table:impliedNormal}
\end{table}
\end{center}
\begin{center}
\begin{table}[h!]
{\begin{tabular}{|l|c|c|c|c|c|}\hline
  Noise/Regularization & 0 & $10^{-10}$ & .01 & .05 & .1 \\\hline
  0\% &$-4.$&$3.999$&&&\\\hline
  5\% &$40.10$&$3.0307$&$3.301$&$4.410$&$5.6205$\\\hline
  10\% &$-4935$&$3.295$&$3.3$&$4.412$&$5.6385$\\\hline
  50\% &$-0.9671$&$-10.52$&$3.359$&$4.513$&$5.764$\\\hline
\end{tabular}}
  \vspace{2pt}
  \caption{Implied curvature at the origin for the graph experiment.}
  \label{table:impliedCurvature}
\end{table}
\end{center}

\subsection{A Sphere} Next we consider an example where $d=3$, starting with the unit
sphere. We take a random sample of points of size $m=80$
\begin{figure}
  \includegraphics[scale=.5]{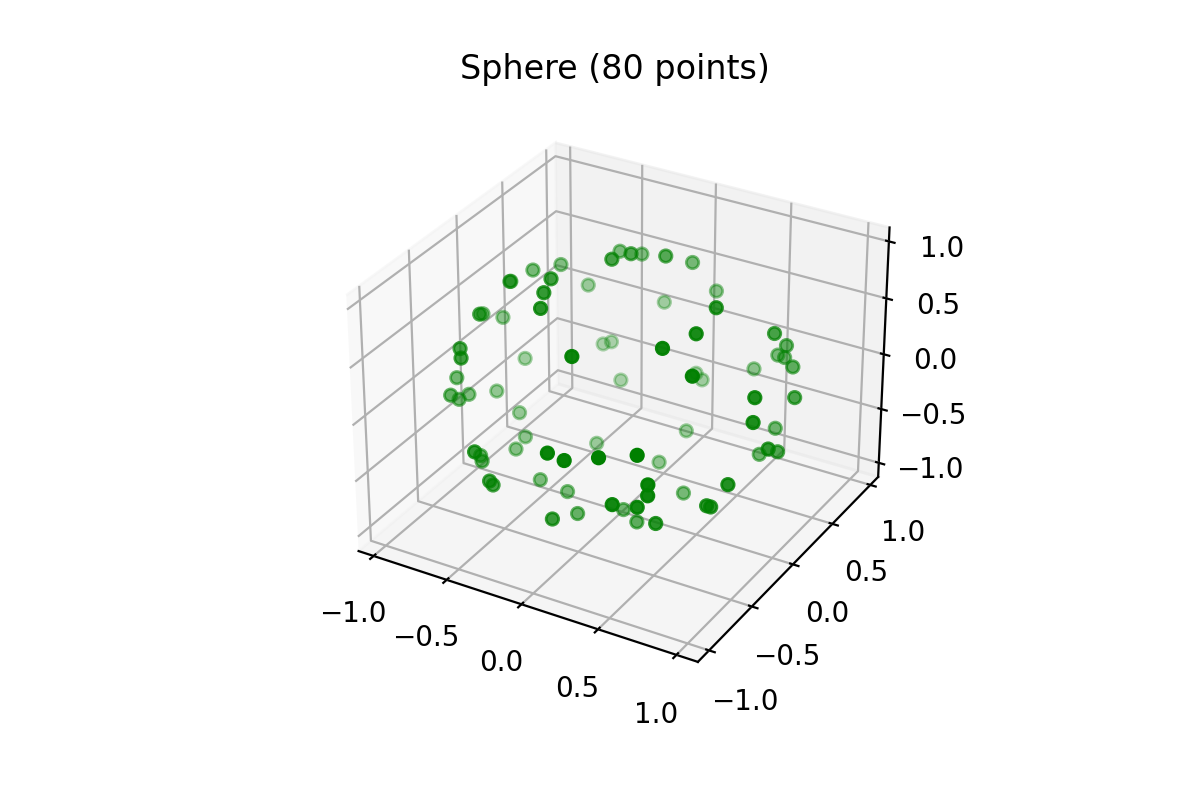}
  \caption{80 points randomly sampled from the unit sphere.}
  \label{fig:sampledSphere}
\end{figure}
that lie exactly on the sphere and used them to compute the signature function,
which, in turn, we use to obtain the implied curvature at another set
of randomly chosen points on the sphere. The random sample is
generated by choosing $x_3$ uniformly at random from $[-1,1]$ and then
setting $x=(\sqrt{1-x_3^2}\cos(\theta),
\sqrt{1-x_3^2}\sin(\theta),x^3)$ for $\theta$ uniformly distributed in
$[0,2\pi)$. It is depicted in Figure \ref{fig:sampledSphere}.
Then we evaluate the signature function and compute the implied
normals and implied curvatures at 32 other points on the sphere
sampled in the same way. The maximum 
deviation from the value 1 of the signature function at these points
is $.0000225$, the maximum angle between the exact normal and the
implied normal is $0.00873$ degrees, and the maximum error in
curvature is $0.0143$.
\subsection{A Folded Shape} 
\begin{figure}
  \includegraphics[scale=.5]{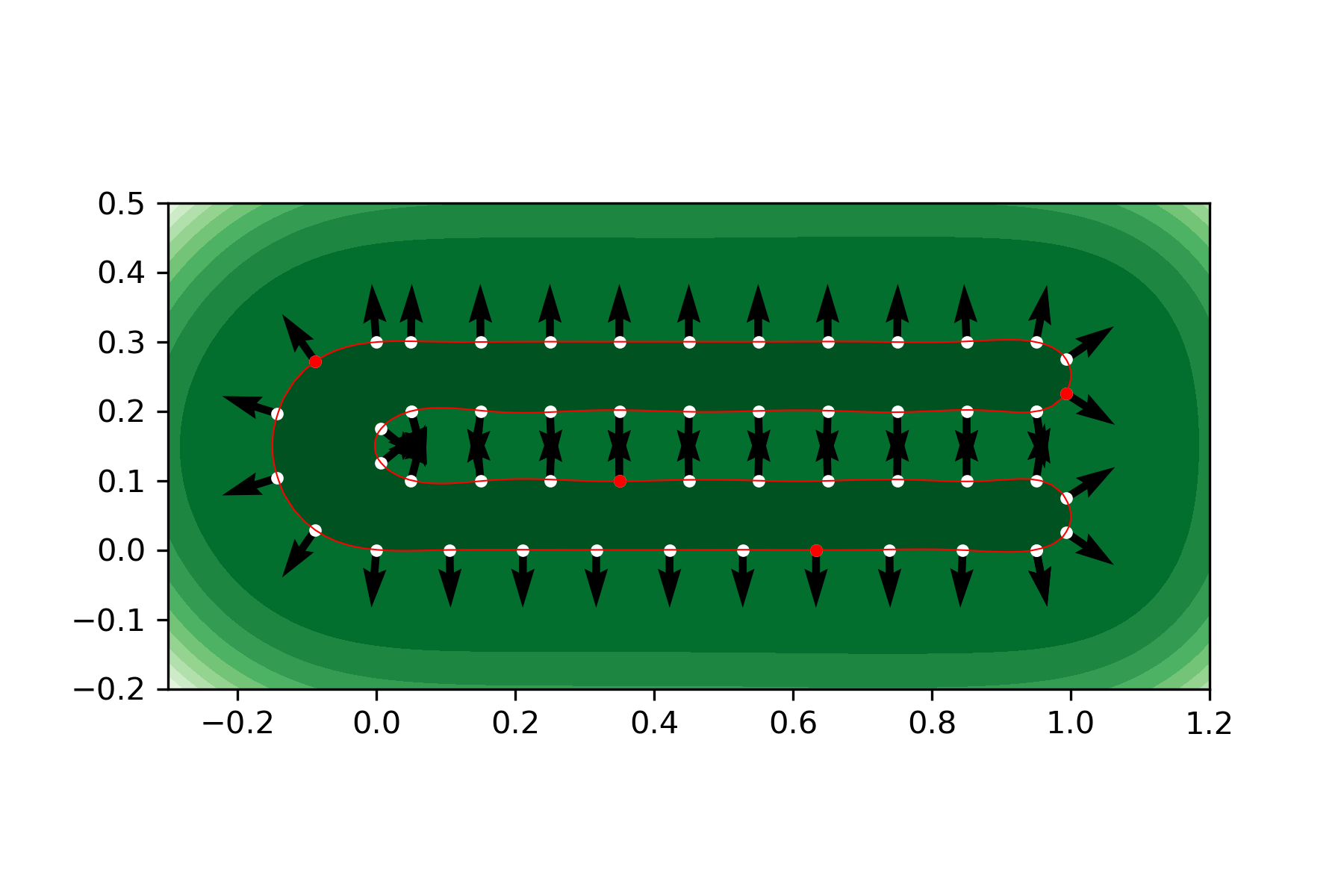}
  \includegraphics[scale=.5]{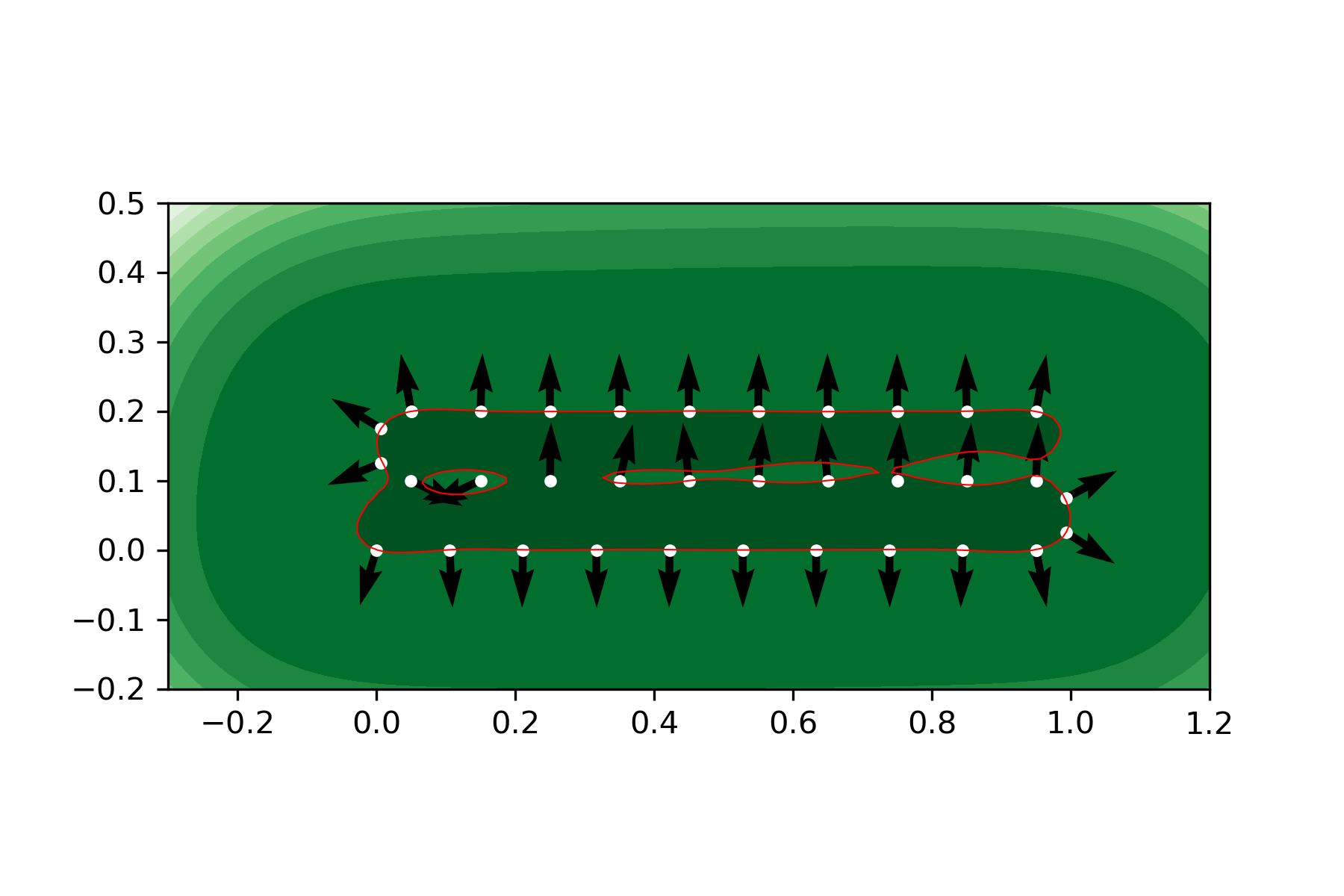}
  \caption{This example illustrates how the method can connect the
    dots properly for closed manifolds (left) even when other methods based
    on local neighborhoods struggle due to limited sample density. This is
    not true for manifolds with boundary (right), at least when the
    sampling density is insufficient. Some of the points in the left cloud are red for
  reasons related to the experiment discussed in the next subsection.}
  \label{fig:sShape}
\end{figure}

Finally we consider an example where
approaches to manifold learning 
or reconstruction via local neighborhoods could lead to erroneous
results unless a dense enough sample is used for the manifold along
with a carefully chosen neighborhood size. We
consider a closed curve obtained as the combination of segments and
semicircles discretized at a level where nearby points in the ambient
space can be far from each other if the distance is measured on the
manifold. The shape is the one implied by its discrete sample in
Figure \ref{fig:sShape} (left) and its accurate interpolation (red line). The
method is clearly able to connect the dots in the proper way. Notice
that, while this is true for closed compact manifolds, this is no longer the
case if the manifold has boundary as Figure \ref{fig:sShape}
(right) clearly shows (at least at this level of discretization). The
curve on the right is obtained by removing the obvious 
part of the closed curve. In this example we used the regularized
($r=1$) Laplace kernel with $\alpha=10^{-10}$.
\subsection{Dimension Estimation}
We use the closed curve $C$ above and the surface $S$ obtained by
replacing each point $(x,0,z)$ on the curve with the segment
$\{x\}\times[-.5,.5]\times\{z\}$. While $S$ is a surface with
boundary, we shall take samples where $y=0$, i.e. away from the
boundary. The idea is to exploit the fact that the signature function
of the curve $C$ viewed as a subset of $\mathbb{R}^3$ contained in
$\mathbb{R}\times\{0\}\times \mathbb{R}$ reproduces this curve as a
limit of the two dimensional surface $[u_C=1-\varepsilon]$ which wraps
around the curve. The sections of this surfaces in the orthogonal
plane to any point along the curve are small ``circles'' (circles
exactly if the curve were a straight line as shown at the end of
Section 4). As a consequence, the
normal to these surfaces turns very quickly in the vicinity of points
on the curve. On the other hand the normal to the surface $S$ does not turn
locally (in a neighborhood the size of which depends on the local curvature)
nearly as much since $S$ is well approximated by its tangent
plane that is also slowly varying across nearby level sets. We therefore take
a sample of points in the immediate vicinity 
of the point of interest (from the data set) and use the
corresponding signature functions  $u_C$ and $u_S$ to compute the
implied normals (to the associated level set surfaces) at these
points. The (numerical) rank $r$ of the  
matrix obtained by using the computed normals as columns provides a
way to estimate the local dimension. The dimension of the tangent
space to the implied hypersurface is cleary $d-1=2$ in both cases but, for
directions in the orthogonal space of the actual manifold, the quick turning
will generate linearly independent vectors, while in tangential
directions, the normals will point in similar directions. This works
for submanifolds of any dimension and  produces an estimate for the
dimension given by $d-r$.

Table \ref{dimEst} below shows the singular values obtained along the central
curve $C$ (corresponding to the red data points in the two dimensional
projection of Figure \ref{fig:sShape}) based on $u_C$ and on $u_S$. The points
are ordered left to right, bottom to top. The first set is for the
curve, whereas the second is for the surface. Shown are
the singular values of a $3\times 15$ matrix generated by taking the
implied normals computed at 15 random perturbations of the points
as its columns. The perturbation consists in translating the point in a
random direction to a point at a random distance uniformly distributed
in $[0,0.01]$. The numerical rank is clearly 2 for the points on the
curve, whereas it is 1 for same points on the surface. Thus the
numerically estimated local dimension at the chosen points is 1 for
the curve and 2 for the surface.
{\small\begin{table}[!htbp]
\centering
\begin{tabular}{|c||c|c|c||c|c|c|}
\hline
 & sv1 & sv2 & sv3 & sv1 & sv2 & sv3 \\
\hline
Point 1 & $3.25$ & $2.11$ & $8.78 \times 10^{-4}$ & $3.87 $ & $1.46 \times 10^{-2}$ & $4.02 \times 10^{-4}$ \\
\hline
Point 2 & $3.42$ & $1.81$ & $2.45 \times 10^{-3}$ & $3.87$ & $5.34 \times 10^{-3}$ & $4.44 \times 10^{-3}$ \\
\hline
Point 3 & $3.28 $ & $2.05$ & $8.39 \times 10^{-2}$ & $3.87 $ & $2.18 \times 10^{-1}$ & $1.66 \times 10^{-3}$ \\
\hline
Point 4 & $3.03$ & $2.41$ & $8.62 \times 10^{-2}$ & $3.87$ & $9.94 \times 10^{-2}$ & $9.07 \times 10^{-5}$ \\
\hline
\end{tabular}
\caption{Estimation of the local dimension based on the rank of the
  span of implied normals at randomly chosen points in the immediate vicinity
  of selected data points.}
\label{dimEst}
\end{table}}
\bibliography{lite.bib} 
\end{document}